\documentclass[10pt]{article}
\usepackage{relsize}

\usepackage{cite}
\usepackage{amsmath,amssymb,amsfonts}
\usepackage{listings}
\usepackage[dvipsnames]{xcolor}
\usepackage{graphicx}
\usepackage{bbm}
\usepackage{amsthm}
\usepackage{color, colortbl}

\definecolor{Green}{rgb}{0.5,1,0.5}
\definecolor{Red}{rgb}{1,0.5,0.5}

\usepackage{textcomp}
\usepackage{arxiv}
\usepackage{url}
\usepackage{mathtools}

\usepackage{textcomp}

\usepackage{cancel}
\usepackage{soul}

\long\def\comment#1{}

\usepackage{caption}

\usepackage{graphicx}%
\usepackage{multirow}%
\usepackage{amsmath,amssymb,amsfonts}%
\usepackage{amsthm}%
\usepackage{mathrsfs}%
\usepackage[title]{appendix}%
\usepackage{xcolor}%
\usepackage{textcomp}%
\usepackage{manyfoot}%
\usepackage{booktabs}%
\usepackage{algorithm}%
\usepackage{algorithmicx}%
\usepackage{algpseudocode}%
\usepackage{listings}%
\usepackage{bm}

\usepackage{rh_defs_21}

\newcommand{\iter}{t}
\newcommand{\stepsize}{\eta}

\def\BibTeX{{\rm B\kern-.05em{\sc i\kern-.025em b}\kern-.08em
    T\kern-.1667em\lower.7ex\hbox{E}\kern-.125emX}}

\usepackage{fancyhdr}
\let\oldsubsection\subsection

\renewcommand{\subsection}[1]{\oldsubsection{\large #1}}

\let\oldsubsubsection\subsubsection

\renewcommand{\subsubsection}[1]{\oldsubsubsection{\large #1}}

\usepackage{authblk}

\author[1]{Reinhard Heckel}
\author[2]{Mathews Jacob}
\author[3]{Akshay Chaudhari}
\author[4,5]{Or Perlman}
\author[6*]{Efrat Shimron}

\affil[1]{Technical University of Munich, Department of Computer Engineering, Munich, Germany %\authorcr Email: {\tt reinhard.heckel@tum.de}
}
\affil[2]{University of Iowa, Department of Electrical and Computer Engineering, Iowa City, IA, USA %\authorcr Email: {\tt mathews-jacob@uiowa.edu}
}
\affil[3]{Stanford University, Department of Radiology and Department of Biomedical Data Science, Stanford, CA, USA %\authorcr Email: {\tt akshaysc@stanford.edu}
}
\affil[4]{Tel Aviv University, Department of Biomedical Engineering, Tel Aviv, Israel %\authorcr Email: {\tt orperlman@tauex.tau.ac.il}
}
\affil[5]{Tel Aviv University, Sagol School of Neuroscience, Tel Aviv, Israel}
\affil[6]{Technion - Israel Institute of Technology, Department of Electrical and Computer Engineering and Department of Biomedical Engineering, Haifa, Israel \authorcr *Email: {\tt efrat.s@technion.ac.il}}
\begin{document}

\title{Deep Learning for Accelerated and Robust MRI Reconstruction: a Review}

\maketitle

\begin{abstract}
   {\large % set font size for the abstract. don't close the brackets here.
    {Deep learning (DL) has recently emerged as a pivotal technology for enhancing magnetic resonance imaging (MRI), a critical tool in diagnostic radiology. This review paper provides a comprehensive overview of recent advances in DL for MRI reconstruction. It focuses on DL approaches and architectures designed to improve image quality, accelerate scans, and address data-related challenges. These include end-to-end neural networks, pre-trained networks, generative models, and self-supervised methods. The paper also discusses the role of DL in optimizing acquisition protocols, enhancing robustness against distribution shifts, and tackling subtle bias. Drawing on the extensive literature and practical insights, it outlines current successes, limitations, and future directions for leveraging DL in MRI reconstruction, while emphasizing the potential of DL to significantly impact clinical imaging practices.}
    }
\end{abstract}


\larger[1] % attempt to make font larger across the entire document

\section{Introduction } 
Magnetic Resonance Imaging (MRI) has long been at the forefront of medical imaging, in that it offers unparalleled ability to visualize the human body's internal structures and functions. Because it is noninvasive and has exceptional soft tissue contrast, it has become an indispensable tool in modern diagnostic medicine. From detecting subtle pathological changes to guiding therapeutic interventions, MRI's versatility is unmatched. However, the full potential of MRI is often constrained by inherent limitations in imaging speed and resolution, which are crucial for accurate diagnosis and patient comfort \cite{constantine2004role}.

One of the fundamental challenges facing MRI is the inherent trade-off between image quality and acquisition time. High-resolution images, which are essential for precise diagnosis, traditionally require longer scan times, which can be burdensome for patients and increase susceptibility to motion artifacts. This limitation is particularly pronounced in dynamic imaging scenarios such as cardiac and abdominal imaging, where rapid physiological movements can cause artifacts such as blurring \cite{Zaitsev2015}. 

Significant research efforts have been aimed at accelerating MRI. Central to these endeavors is the development of methods for image reconstruction from under-sampled data. Techniques based on parallel imaging (PI) introduced in the 1990s were an important watershed \cite{Sodickson1997,griswold2002generalized,Pruessmann1999,Seiberlich2007,Deshmane2015parallel,lustig2010spirit}. These techniques leverage the spatial diversity of multiple coil arrays to reconstruct images, thus allowing for reduced scan times by acquiring less data. The early 2000s saw the emergence of compressed sensing (CS) methods, which constituted a novel approach to MRI reconstruction \cite{lustig2007sparse,donoho2006compressed,jung2009k,feng2016xd,eldar2011CS,ye2019compressed,jacobs013blind,fessler2020optimization}. CS overcomes the sparsity of MRI images by enabling the reconstruction of high-quality images from a much smaller set of measurements than traditionally required. The development of these techniques represented major milestones in MRI, and contributed to substantial reductions in scan duration and improvements in image quality. However, both parallel imaging and compressed sensing methods have their limitations, such as the need for specific acquisition schemes or the long computation times required for solving optimization problems, especially for non-Cartesian imaging trajectories. 

In recent years, the advent of deep learning (DL) has heralded a new era in MRI by offering promising solutions to these longstanding challenges. DL, a subset of machine learning characterized by algorithms based on computational neural networks, has had remarkable success in extracting complex patterns from large datasets 
 \cite{lecun2015deep,alzubaidi2021review,shrestha2019review,shakes2021review}.In the realm of MR image reconstruction, DL methods focus on learning from vast amounts of data to transform under-sampled or noisy data into high-fidelity images.These methods have demonstrated their ability to mitigate artifacts, enhance resolution, and accelerate the imaging process \cite{zhu2018image,ShanshanWang2020,hammernik2018MRI,aggarwal2018modl,lundervold2019overview,mazurowski2019deep,ravishankar2019image,knoll2020deep,liang2020deep}. There are several datasets \cite{knoll2020fastmri,ong2018mridata,souza2018open,desai2021skm} and community challenges \cite{muckley2021results,knoll2020advancing,muckley2020state,tolpadi2023k2s} for MRI reconstruction research. 

This review provides a comprehensive overview of recent advances and applications of deep learning (DL) to the reconstruction of magnetic resonance (MR) images. Given the rapid pace at which this field is evolving, encapsulating the entirety of the published literature is a formidable challenge. Previous reviews have laid the groundwork by detailing the fundamental components of DL architectures \cite{lundervold2019overview,mazurowski2019deep,ravishankar2019image,knoll2020deep,liang2020deep,pal2022review,lin2021artificial,Oscanoa2023,Spieker2023}. Here, we cover the diversity of approaches, by spotlighting not only the emergence of novel methods such as self-supervised learning methods and diffusion models, but also by addressing areas where DL encounters significant hurdles, including susceptibility to distribution shifts, instabilities, and inherent biases. Drawing on our hands-on experience, we propose actionable strategies to enhance the robustness of DL models to mitigate these challenges effectively.

%\vspace{-15mm}

\section{Background on MRI reconstruction}

%In this section, we describe the image formation process and forward model, and discuss conventional, optimization-based, non-DL image reconstruction methods.

\subsection{Image formation \& forward model}

The acquisition process in many imaging schemes can be modeled by an operator  $\mathbf A$ applied on the continuous domain image $\mathbf x$, where the process of collecting measurements is described by $\mathbf y=\mathbf A(\vx) + \mathbf n$. 
In MRI, the measurement operator $\mathbf A$ commonly corresponds to a multi-coil Fourier sampling operator. Although the acquisition is continuous, the general practice is to discretize the problem. Thus, we consider the reconstruction of an image vector $\mathbf x$ from linear measurements, modeled by a matrix $\mathbf A$, by:
\begin{equation}
\label{forwardmodel}
\mathbf y=\mathbf A\mathbf x + \mathbf n.
\end{equation}
The above equation is a numerical model for the imaging device, and is often referred to as the \emph{forward model}. In many imaging methods, the forward model is known precisely. However, there are many applications where the forward model is unknown or only partially known. Examples include imaging in the presence of motion during the acquisition, trajectory errors, and field inhomogeneity effects in MRI acquisitions.

Due to MRI's long scan duration, many scans are accelerated by sampling \emph{k-space} at a sub-Nyquist rate. In these cases, the forward model  $\mathbf A$ is often rank-deficient, making the recovery of $\mathbf x$ an ill-posed problem. 

\subsubsection{Conventional model-based image recovery}

When the recovery of $\mathbf x$ is ill-posed, many MRI schemes such as SENSE \cite{Pruessmann1999} pose the recovery as an optimization problem $\mathbf x = \arg \min_{\mathbf x} \mathcal L(\mathbf x)$ with an objective function  
\begin{equation}
\label{modelpast}
\mathcal L(\mathbf x)=\underbrace{\|\mathbf A\,\mathbf x-\mathbf y\|_2^2}_{\mbox{data
		consistency}} + {\lambda}~\underbrace{\mathcal R(\mathbf x)}_{\mbox{regularization}}
\end{equation}
where the first term is often called a \emph{data consistency} term, and the second term is called a \emph{regularization} prior. The objective \eqref{modelpast} is sometimes called a \emph{variational objective} \cite{hammernik2018learning}. 

The prior $\mathcal R: \mathbb C^n \rightarrow \mathbb R_{+}$ is used to restrict the solutions to the space of desirable images. The prior $\mathcal R(\mathbf x)$ has a large value when $\mathbf x$ is an undesirable image and is small for a desirable image. A common prior used in compressive sensing methods is wavelet-domain sparsity, where the number of non-zero wavelet coefficients or their surrogates are used as priors \cite{lustig2007sparse,donoho2006compressed}. In this case, the optimization algorithm facilitates the recovery of an image $\mathbf x$ that has few non-zero wavelet coefficients.

From a Bayesian perspective, the above formulation can be viewed as an a-posterior estimate \cite{ravishankar2019image,bilgic2011multi}, where the goal is to find an image $\mathbf x$ that maximizes the posterior distribution $p(\mathbf x|\mathbf y) = \frac{p(\mathbf y|\mathbf x)p(\mathbf x)}{p(\mathbf y)}$. The estimate is obtained by minimizing the negative log posterior 

\begin{equation}
-\log p(\mathbf x|\mathbf y) = \underbrace{-\log p(\mathbf y|\mathbf x)}_{\mbox{data
		consistency}} - \underbrace{\log p(\mathbf x)}_{\mbox{prior}}.
\end{equation}

Here, the first term is the data consistency term. It yields the mean-squared error in equation \eqref{forwardmodel} if the noise vector $\mathbf n$ has i.i.d. Gaussian entries. Data consistency terms appear in both compressed sensing \cite{lustig2007sparse} and DL methods \cite{hammernik2021systematic}, as they ensure that the reconstructed images adhere closely to the acquired data. The second term incorporates prior information on the images \cite{luo2020mri,bilgic2011multi}. 

Over the past few decades, substantial research efforts have been dedicated to crafting effective priors. Tikhonov regularization, for instance, employs a Gaussian prior on $\mathbf x$, resulting in a regularization term $\mathcal R(\mathbf x) =\|\mathbf x\|^2$, and compressed sensing methods mentioned earlier promote sparsity. 

\subsubsection{Optimization algorithms}
\label{algorithms}

The loss in equation~\eqref{modelpast} is typically minimized by applying  iterative first-order optimization algorithms such as gradient descent. 
Starting from $\vx_0$ with a stepsize of $\stepsize_\iter$, iteration $t+1$ of gradient descent is described by
\begin{align}
\vx_{\iter+1}
&=
\vx_\iter - \stepsize_\iter \mc L(\vx_\iter) \label{eq:gditeratesvar} \\
&=
\vx_\iter - \stepsize_\iter
\left(
\transp{\mA} (\mA\vx_\iter - \vy)
+ 
\nabla \mathcal R(\vx_\iter) \right). \nonumber
\end{align}
The above algorithm depends on the gradient of the regularizer $\nabla \mathcal R$. When $\mathcal R(\mathbf x) = \log p(\mathbf x)$ is the log-prior as mentioned earlier, $\nabla R(\mathbf x) = \nabla \log p(\mathbf x)$, which is often referred to as the \emph{score} of the distribution. This term can be seen as an estimate of the noise. Subtracting it from the original image will give a cleaner image. Hence, the above gradient descent can be seen as a noise removal step, where the alias components specified by $\transp{\mA} (\mA\vx_\iter - \vy)$ and the noise components specified by $\nabla R(\mathbf x)$ are removed from the current estimate $\vx_{\iter}$. 

Other popular fast iterative algorithms for minimizing the objective~\eqref{modelpast} include the alternating direction method of multipliers (ADMM) \cite{boyd2011distributed} and the fast iterative shrinkage thresholding algorithm (FISTA) \cite{beck2009fast}. For example, the ADMM scheme considers the equivalent problem, 
\begin{equation}
\mathbf x = \arg \min_{\mathbf
	x} \min_{\mathbf v} \|\mathbf A\,\mathbf x-\mathbf y\|_2^2 + {\lambda}~\mathcal R(\mathbf v) ~ \mbox{such that}~~ \mathbf v=\mathbf x
\end{equation}
The above problem is solved by alternating between the following steps 
\begin{eqnarray}\label{data}
{\mathbf x}_{n+1} &=&  \arg \min_{\mathbf x}  \|\mathbf A\,\mathbf x-\mathbf y\|_2^2 + \beta\|\mathbf x-(\mathbf v_n-\mathbf u_n)\|^2 \\
{\mathbf v}_{n+1} &=& \beta\|\mathbf v_n-\underbrace{\left(\mathbf x_{n+1}-\mathbf u_n\right)}_{\overline{\mathbf x}}\|^2 + \lambda R(\mathbf v) \\
\label{update}
\mathbf u_{n+1} &=& \mathbf u_{n} + (\widehat{\mathbf x_{n+1}} - \widehat{\mathbf v}_n)
\end{eqnarray}
The second step of the above optimization scheme
\begin{eqnarray}\label{denoising2}
{\mathbf v} &=& \arg \min_{\mathbf v}   \beta\|\mathbf v-\overline{\mathbf x}\|^2 + \lambda \mathcal R(\mathbf v) = \mathcal D_{\beta}(\overline{\mathbf x})
\end{eqnarray}
can be viewed as a \emph{denoising} step to clean the current solution $\overline{\mathbf x}$, thus yielding $\mathbf v$. For many penalties (e.g. $\ell_1$ norm), the solution to \eqref{denoising2} can be evaluated as proximal mapping. Here, $\beta$ is a continuation parameter that can be interpreted as $\frac{1}{\sigma_{\beta}^2}$, where $\sigma_{\beta}$ is the variance of \emph{noise} in $\overline{\mathbf x}$ that decreases with the iterations. The first step \eqref{data} involves an inversion step to reduce the cost function composed of the linear combination of the data consistency error and the deviation from the \emph{denoised} image $\mathbf v$.  This provides an iterative denoising interpretation, which is used in plug-and-play algorithms (discussed below, in Section \ref{secplug}). One challenge with the above convex optimization schemes is their high computational complexity, which is due to the numerous iterations required for convergence. In particular, the data consistency step \eqref{data} involves the evaluation of the forward model and its adjoint, which is often computationally expensive.

% =============== main section ===================
\section{DL reconstruction: approaches and architectures}

There are at least four different approaches to deep learning based MRI: (i) neural networks trained end-to-end; (ii) approaches based on pre-trained denoisers, often called plug-and-play methods;
(iii) approaches based on generative models; 
and (iv) un-trained neural networks. In this section, we review these four approaches. It is worth mentioning that in the 2020 FastMRI challenge, the top-performing methods were all neural networks trained end-to-end \cite{Muckley2021}. However, relative comparisons of algorithms depend on the problem setup and metrics. In addition, the field is progressing rapidly, so that new and diverse benchmarking studies would be valuable. 

\subsection{Neural networks trained end-to-end } %\blue{(Efrat and Reinhard)}
\label{sec:end-to-end}

Neural networks trained end-to-end are commonly trained to map the acquired data, which is often noisy and degraded by undersampling artifacts, to a target, ground-truth image. Their training hence requires such paired data. 

Let $f_\theta$ be a neural network, which receives the measurements as input and produces clean, reconstructed images as output. Given a training set consisting of pairs of measurements and target images $\{(\vy_1,\vx_1),\ldots,(\vy_n,\vx_n)\}$, the network is trained by minimizing the loss between the prediction of the network and the target images, i.e.,

\begin{align} 
\label{eq:end-to-end-training}
\mc L(\vth) = \sum_{i=1}^n \text{loss}( f_\theta(\vy_i), \vx_i ).
\end{align}

%%%
\subsubsection{Networks mapping a noisy measurement to a clean image}

Many architectures have been developed for mapping a noisy measurement to a clean image. Here we review some of the most well-known approaches.

One of the early architectures is known as  AUTOMAP (Automated Transform by Manifold Approximation) \cite{zhu2018image}. It utilizes a fully connected network followed by a convolutional network as the  architecture $f_\theta$ in equation (\ref{eq:end-to-end-training}). This architecture does not incorporate the known forward model directly. Instead, it maps data from \emph{k-space} to image domain, and learns the forward model from the data.

Other architectures commonly utilize image-to-image neural networks and incorporate the forward map. 
This class of approaches maps a coarse reconstruction of the image, for example a zero-filled image, to a target image. In the notation above, the architecture $f_\theta$ consists of a linear map computing the zero-filled image followed by application of an image-to-image neural network.

The architecture of the image-to-image network is most commonly a convolutional neural network (CNNs). 
One of the pioneering works in this area was by Wang et al. \cite{Wang2016}, which demonstrated that using a CNN as the image-to-image network enables substantial improvements in both speed and image quality. 
Another pioneering work was by \cite{unser2017deep}, which showed that this approach is applicable to a wide range of linear forward models.
% \cite{lee2018deep}.

Additionally, numerous other studies have embedded UNET, ResNet or recurrent neural networks as the backbone architecture \cite{lundervold2019overview,mazurowski2019deep,ravishankar2019image,knoll2020deep,liang2020deep,pal2022review,lin2021artificial,Oscanoa2023,Spieker2023}. More recently, vision transformers have been utilized instead of a CNN as the image-to-image network. Several studies demonstrated that transformers can provide  improvements~\cite{Lin2022,lin2022vision,guo2023reconformer}. However, they are computationally more expensive. In practice, a UNET provides a very good trade-off between image quality and computational performance.

\subsubsection{Unrolled network architectures}
\label{sec_unrlled}

Currently, some of the best-performing neural networks are based on 
unrolled architectures \cite{Muckley2021}. These networks are obtained by unrolling an iterative algorithm such as gradient descent. The idea of unrolled networks was first introduced by \cite{gregor_LearningFastApproximations_2010_v2}, and several pioneering works applied it in the context of MRI reconstruction ~\cite{yang_DeepADMMNetCompressive_2016_v2,hammernik2018MRI,Schlemper2018,aggarwal2018modl}. These architectures iterate between two types of blocks: (i) data-consistency blocks, which can be computed using different algorithms \cite{hammernik2021systematic}, and (ii) blocks that remove noise and artifacts, which are commonly implemented by a deep neural network. 

One of the early works in this context was by Hammernik et al. ~\cite{hammernik2018MRI}, who introduced the \emph{variational network}. This approach relies on a gradient descent algorithm to minimize the variational objective~\eqref{modelpast} where the regularizer $\mc R$ is taken as the total-variation norm. In this case, the gradient of the regularizer in the gradient descent iterations~\eqref{eq:gditeratesvar} takes the form of a convolution. Thus, the gradient descent iterations can be interpreted as a neural network that applies data consistency operations (originating from the gradient of the least-squares loss) and the application of a convolutional network (originating from the gradient of the regularizer). Motivated by this observation, these so-called variational networks initialize $\vx_0=\transp{\mA}\vy$ and then perform the following computations with a neural network:
\begin{align}
\label{eq:itervarnetv1}
\vx_{\iter+1}
=
\vx_\iter - \stepsize_\iter
\transp{\mA} (\mA\vx_\iter - \vy)
+ 
\mathrm{CNN}_\iter(\vx_\iter).
\end{align}
Here, both the parameter $\stepsize_\iter$ and the parameters of the CNN are learnable. The original variational network~\cite{hammernik2018MRI} used a relatively shallow CNN, inspired by the parameterization provided by the total variation norm and its generalizations; specifically, the fields-of-experts-model. This yields a well-performing network with very few parameters. Later work has also shown that using a UNET within the unrolled network can improve the overall performance.  \cite{aggarwal2018modl,sriram_EndtoEndVariationalNetworks_2020_v2}.

Other unrolled methods adopted alternate algorithms to minimize the variational loss~\eqref{eq:gditeratesvar} (described in Section \ref{algorithms}). Those replaced the CNN in equation~\eqref{eq:itervarnetv1} with other image-to-image architectures; see~\cite{yang_DeepADMMNetCompressive_2016_v2,Schlemper2018,aggarwal2018modl,fabian_HUMUSNetHybridUnrolled_2022_v2,darestani_IRFRestormerIterativeRefinement_2024_v2} for a few examples. 
Furthermore, other architectures replaced the image-domain CNN with either a \emph{k-space} CNN, a dual-domain (\emph{k-space} and image domain) network \cite{deepslr,eo2018kiki,wang2024dct} or a transformer \cite{fabian_HUMUSNetHybridUnrolled_2022_v2}. 

\emph{\textbf{Computational considerations.}} 
The unrolling step requires multiple physical realizations of the CNN block during training, which translates into a high memory demand during training. This restricts their application in higher dimensional (e.g. 3D, 4D) applications. Programming solutions such as gradient check-pointing are now available to reduce the memory demand, at the expense of increased computational complexity. An alternative approach relies on deep equilibrium models  \cite{deq,mol}. These models use a single CNN block and iterate the steps \eqref{data}-\eqref{update} until convergence to a fixed point, similar to PnP methods. These methods then implement the fixed point iterations for back propagation. These methods thus enable the evaluation of the forward and back propagation using a single physical CNN block, thus reducing the memory demand. The MOL \cite{pramanik2023accelerated} method also imposes a local Lipschitz constraint on the CNN block, which offers theoretical guarantees and robustness without sacrificing performance.

\subsection{Pretrained plug-and-play (PnP) methods}
\label{secplug}

Early CS methods relied on convex priors $\mathcal R(\mathbf x)$, such as the total-variation norm. 
Plug-and-play (PnP) methods make it possible to solve inverse problems with pre-trained denoisers. Another benefit is that they work with arbitrary forward models, where the prior incorporates information about the image.  One class of PnP methods replaces the proximal operator $\mathcal D_{\beta}$ in \eqref{denoising2} with a pre-trained denoiser. 
While early methods relied on off-the-shelf image denoisers such as BM3D \cite{bm3d},  %assuming an input noise variance of $\sigma_{\beta}$, 
pre-trained CNN denoisers are now considered to be more effective \cite{ahmad2020plug,kamilov_PlugandPlayMethodsIntegrating_2022,ryu2019plug}. Note that the proximal step in \eqref{denoising2}
\begin{eqnarray}
\mathcal D_{1/2\sigma^2}({\mathbf x}) &=& \arg \min_{\mathbf v}  \frac{1}{2\sigma^2}\|\mathbf v-\mathbf x\|^2 + \lambda~ \mathcal R(\mathbf v) 
\end{eqnarray}
can be seen as the maximum a-posteriori (MAP) estimate of $\mathbf x$ from its noise corrupted measurements
\begin{equation}
\mathbf v = \mathbf x + \sigma ~\mathbf n.
\end{equation}
Here, $\mathbf n \sim \mathcal N(0,\mathbf I)$ is a sample from a Gaussian distribution with variance $\sigma^2$. The CNN modules are hence pre-learned from training data as MAP denoisers, where noise- corrupted images are fed as input and the model is trained to yield noise-free images. During inference, steps \eqref{data}-\eqref{update} are iterated until the algorithm converges to a fixed point. Similar to CS methods, several iterations are often needed for convergence, which translates into higher computational complexity than the unrolled approaches described in section \ref{secplug}.

\subsection{Generative priors}

Another successful approach to DL-based MRI reconstruction is to learn an image prior parameterized by a generative neural network. Several major classes of generative methods have emerged, based on variational autoencoders \cite{kingma2013auto,tezcan2018mr}, Generative Adversarial Networks (GANs) \cite{goodfellow2014generative,yang2017dagan,mardani2018deep_v2,quan2018compressed,cole2020unsupervised,lv2021gan,korkmaz2022unsupervised}, and very recently, diffusion models \cite{yang2023diffusion,po2023state,croitoru2023diffusion,kawar_DenoisingDiffusionRestoration_2022_v2,bora_CompressedSensingUsing_2017_v2,asim_InvertibleGenerativeModels_2020_v2,jalal_RobustCompressedSensing_2021_v2,daras_SolvingInverseProblems_2021_v2,gungor2023adaptive,ho2020denoising}. Here we focus on the two latter ones, which have attracted substantial attention. 

One of the major advantages of generative approaches for image reconstruction is that they are flexible with regard to changes of the forward model, and at the same time perform well for reconstructing high-quality images from undersampled data. Furthermore, their probabilistic nature  provides measures for uncertainty quantification, which is highly important for clinical imaging \cite{chung2022score,luo2023bayesian,zach2023stable}.

\subsubsection{GANs} % \blue{(Efrat)

GANs~\cite{goodfellow2014generative} are a framework for generative modeling. 
A GAN consists of two competing neural networks: a generator, which aims to produce data indistinguishable from a given dataset of real images, and a discriminator, whose role is to distinguish between the generator's output and the real data. GANs are trained using an  adversarial loss ~\cite{goodfellow2014generative}; this process enables the generator to learn to generate high-quality realistic images. After training, the generator can be used either to generate images that look similar to those in the training set, or as a prior for image reconstruction. 

In the context of MRI reconstruction, GANs have attracted substantial attention over the last few years \cite{yang2017dagan,mardani2018deep_v2,quan2018compressed,cole2020unsupervised,lv2021gan,korkmaz2022unsupervised}. For example, DAGAN (Deep De-Aliasing Generative Adversarial Networks) \cite{yang2017dagan} was a pioneering work that proposed a conditional GAN with a refinement-learning stage, and used a loss function comprised of an adversarial and a perceptual component. Mardani et al. \cite{mardani2018deep} proposed a reconstruction framework where GANs were used for learning the low-dimensional manifold that underlies high-quality MR images. However, images generated by the generator are not necessarily consistent with the acquired measurements. To ensure such consistency, they included an affine projection operation, conducted  by a layer placed between the generator and discriminator. Another approach for tackling this was proposed by Quan et al. \cite{quan2018compressed}, who introduced a novel cyclic loss in their GAN architecture to enforce data consistency. These methods, and many others \cite{lv2021gan,ali2022role} showcased the potential of GANs to produce clinically viable MRI reconstructions.

\subsubsection{Diffusion models} %\blue{(Efrat)
\label{diffusion_models_section}
Diffusion models, a class of generative models that have garnered substantial attention in recent years, are making an impact in a variety of  fields, including MRI reconstruction \cite{yang2023diffusion,po2023state,croitoru2023diffusion,
kawar_DenoisingDiffusionRestoration_2022_v2,
bora_CompressedSensingUsing_2017_v2,
asim_InvertibleGenerativeModels_2020_v2,
jalal_RobustCompressedSensing_2021_v2,
daras_SolvingInverseProblems_2021_v2,gungor2023adaptive,
ho2020denoising}. These models operate by learning to reverse a diffusion process that gradually transforms random noise into structured images, and have shown a remarkable capability to generate high-quality, detailed images. 

Diffusion models have been derived using different approaches \cite{yang2023diffusion,po2023state}, including discretized corruptions, e.g., denoising diffusion probabilistic models (DDPMs) \cite{ho2020denoising}, denoising score matching \cite{song_ImprovedTechniquesTraining_2020_v2}, and continuous formulations based on stochastic differential equations (SDEs) \cite{song2020score}. 

For a general probability density function $p(x)$, these approaches approximate the \emph{score function}, defined by $\nabla_{x} \log p(x)$, using a neural network $s_\theta(x)$. To do so, the network is used to approximate a series of conditional score functions, $s_t(x_t) = \nabla_{x_t} \log p(x_t | x_{t+1})$, which guide the denoising process from pure noise, i.e., $x_T$ drawn from a normal Gaussian distribution for some maximum iteration value $T$, to a clean sample $x_0 \sim p(x)$. Once trained, these models can be used to sample unconditionally from the prior distribution by running the reverse diffusion process, and hence generating new samples.

In the context of inverse problems in general, and MRI reconstruction in particular, the diffusion process can be hijacked to approximately sample from the conditional posterior distribution, $p(y|x)$ instead. One method involves conditioning on the \emph{k-space} measurements $y$ and applying  Bayes' rule to the series of score functions, i.e., 
\begin{align}
\label{eq:diffusion_models1}
\nabla{\log p(x_t|y,x_{t+1})} 
= \nabla{\log p(y|x_t)} + \nabla {\log p(x_t|x_{t+1})}.
\end{align}
The second term, corresponding to the prior conditioned on the denoising process, is unchanged from the original diffusion model and can be learned by training on clean, fully sampled images. The first term, corresponding to the likelihood conditioned on the denoising process, can be approximated through various approaches \cite{jalal_RobustCompressedSensing_2021_v2,ho2020denoising,chung2022diffusion}.
In a naive approximation, 
\begin{align}
\label{eq:diffusion_models2}
\nabla {\log p(y|x_t)} \approx  A^H(Ax_t - y),
\end{align}
given the MRI forward model.

A growing body of work demonstrates that diffusion models work well for accelerated MRI and exhibit flexibility when handling various sampling patterns 
\cite{jalal_RobustCompressedSensing_2021_v2,zach2023stable,gungor2023adaptive,chung2022score,daras_SolvingInverseProblems_2021_v2,yu2023universal}.  For example, in a pioneering work, Jalal et al. \cite{Jalal2021} demonstrated that training a score-based generative model using Langevin dynamics, without making any assumptions on the measurement system, could yield competitive reconstruction results for both in-distribution and out-of-distribution data.  Chung et al. \cite{chung2022score} demonstrated that score-based diffusion models trained solely on magnitude images can be utilized for reconstructing complex-valued data.  Luo et al. \cite{luo2023bayesian} described a comprehensive approach using data-driven Markov chains for MRI reconstruction which not only facilitates efficient image reconstruction across variable sampling schemes, but also enables the generation of uncertainty maps. 

The flexibility afforded by explicitly decoupling the image prior (which is learned with diffusion models) and the statistical measurement model has also enabled other extensions. These include incorporating errors into the forward model, e.g., due to motion \cite{levac_AcceleratedMotionCorrection_2023_v2} and field inhomogeneity \cite{alkan2023variational} and incorporating multiple image contrasts \cite{levac2023mri}.

% =============== untrained networks ===========
\subsection{Un-trained neural networks} %}\blue{(Reinhard,Mathews)}}

Un-trained methods are DL models that do not rely on training data apart from hyper-parameter tuning. Instead of conventional training on large datasets, these methods are typically based on fitting a randomly initialized neural network to a specific measurement. Here we discuss two types of methods: un-trained neural networks based on CNNs, and methods based on coordinate-wise implicit neural networks.

\subsubsection{Un-trained CNNs for single image recovery}
CNNs can be used as an image prior by fitting a randomly initialized CNN with gradient descent to a measurement. This approach, termed the deep image prior (DIP), was introduced in a pioneering work by Ulyanov et al. \cite{ulyanov_DeepImagePrior_2020_v2}. The optimization problem is formulated by,
\begin{equation}\label{key}
\mathbf x^* = \arg \min_{\theta} \|\mathbf A\mathbf x-\mathbf y\|^2 ~~\mbox{such that}~~ \mathbf x = \mathcal G_{\theta}(z),
\end{equation}
where $\mathcal G_{\theta}$ is a CNN generator whose input $\mathbf z$ is a noise vector drawn from some noise distribution. The optimization is performed using gradient descent or ADAM \cite{Kingma2015}, starting with random initialization of the network weights, and early stopping is used for regularization. 
The image quality first improves with the number of iterations, and then degrades as the network begins to fit the measurement noise in $\mathbf y$. This behavior is caused by the implicit bias of CNN networks to natural images: when trained with gradient descent, CNNs fit the smooth images before the noise, as formalized in \cite{heckel_DenoisingRegularizationExploiting_2020_v2}. 

Un-trained networks perform very well for denoising~\cite{ulyanov_DeepImagePrior_2020_v2,heckel_DeepDecoderConcise_2019_v2} and compressive sensing (e.g., accelerated MRI). These methods can provably denoise smooth signals~\cite{heckel_DenoisingRegularizationExploiting_2020_v2} and can provably reconstruct undersampled smooth images \cite{heckel_CompressiveSensingUntrained_2020_v2}. Un-trained networks also work quite well for accelerated MRI; they provide significant improvement over sparsity-based methods for 2D accelerated MRI~\cite{Darestani_Heckel_2021}.

One key benefit of un-trained networks is that they do not need training data. However, this benefit comes at the expense of performance; the images produced by DIP methods are commonly not comparable to those from the pre-trained networks discussed above. In addition, DIP often suffers from longer run times compared to the unrolled and direct inversion approaches because of the need for ADAM or gradient descent optimization during reconstruction.

\subsubsection{Un-trained CNNs for joint recovery of multiple images}
\label{tdip}
Recently, the DIP framework was extended to dynamic imaging applications \cite{jin_TimeDependentDeepImage_2019_v2,genStorm} where the images in a time series are modeled as the output of a generator
\begin{equation}\label{SToRModel}
\gamma_{t}(\mathbf r) = \mathcal{G}_{\theta}\left(\mathbf{z}_{t}\right).
\end{equation}
Unlike the fixed noisy input used in the original DIP work~\cite{ulyanov_DeepImagePrior_2020_v2}, here $\mathbf{z}_{t}$ are low dimensional latent vectors at a specific time point $t$. $\mathcal{G}_{\theta}$ is a deep CNN generator, whose weights $\theta$ are independent of $t$. For example, in a free-breathing cardiac MRI, the images in the time series at a specific time $t$ can be viewed as non-linear functions of cardiac and respiratory phases captured by $\mathbf z_t$. This model \eqref{SToRModel} can be viewed as a non-linear mapping/lifting from a low-dimensional subspace $\mathbf Z$ to the image space. The low-dimensional nature of the latent vectors enables the exploitation of the non-local redundancies between images at different time points, thus facilitating the fusion of information between them as in \cite{ahmed2020free,poddar2019manifold}. 

The network parameters $\theta$ and the latent variables $\mathbf{z}$ are jointly optimized for by minimizing the cost function
\begin{equation}\label{proposed}
\mathcal C(\mathbf z,\theta) =\sum_{t=1}^N\|\mathcal A_{t}\left(\mathcal G_{\theta}[\mathbf z_{t}]\right) - \mathbf y_{t}\|^2  + \lambda_1 \underbrace{\|\nabla_{\mathbf z} \mathcal G_{\theta}\|^2}_{\scriptsize \mbox{network regularization}}  + {{\lambda}}_2 \underbrace{\mathcal{R}(\mathbf{z})}_{\scriptsize\mbox{latent regularization}}.
\end{equation}
The network regularization is an $\ell_2$ penalty on the weights $\theta$, which was shown to minimize the need for early stopping and provide improved performance. The latent vector regularization term involves a smoothness regularization to capitalize on the temporal smoothness of the images in the time series. 

The above approach can also be extended to 3D applications, where the joint alignment and recovery of data from different slices obtained using different acquisitions may differ in cardiac/respiratory motion. Different set of latent vectors are used for different slices to account for differences in breathing patterns and cardiac motion. In this case, a Kullback-Leibler divergence term is used to encourage the latent vectors of all the slices to follow a zero-mean Gaussian distribution, thus facilitating the alignment of data from different slices. 

\subsubsection{Coordinate-based networks}

%\textcolor{red}{fix references here - there are too many brackets in the refs}

Coordinate-based neural representations, also known as implicit neural representations (NeRF-type networks), have recently emerged as an efficient way to represent and work with images, 3D shapes, and other signals \cite{mildenhall2021nerf}. They are commonly used for representing scenes and performing view syntheses in vision~\cite{arandjelovic_NeRFDetailLearning_2021_v2,williams_DeepGeometricPrior_2019_v2,huang2023neural}. To represent a 2D or 3D object, these models map a coordinate input (e.g., $(x,y)$-coordinate for 2D or $(x,y,z)$-coordinate for 3D) to a pixel value, e.g., a real number for a gray-scale image and two real numbers for a complex-valued image. 

Coordinate-based networks can be used in an analogous fashion to un-trained CNNs to reconstruct an image \cite{tancik_FourierFeaturesLet_2020a_v2,dupont_COINCOmpressionImplicit_2021_v2}. Specifically, they can replace the CNN in an un-trained network and be fitted to measurement data. Networks with Fourier-feature input (like NerF~\cite{arandjelovic_NeRFDetailLearning_2021_v2}, SIREN~\cite{sitzmann_ImplicitNeuralRepresentations_2020_v2}, and Fourier Feature inputs~\cite{tancik_FourierFeaturesLet_2020a_v2}) impose a smoothness prior similar to the un-trained CNN discussed in the previous section and work reasonably well. However, they perform slightly worse in terms of image quality. In the context of MRI reconstruction, coordinate networks are useful for high-dimensional objects such as 3D volumes and scenarios with motion. For example, \cite{huang_NeuralImplicitKSpace_2022_v2} and \cite{kunz_ImplicitNeuralNetworks_2023_v2} used coordinate networks successfully for free-breathing cardiac MRI reconstruction and achieved good results.

%%%
\subsection{Self-supervised methods} % \blue{(Reinhard, Efrat,Mathews)

Neural networks, such as the end-to-end networks discussed in Section~\ref{sec:end-to-end} are usually trained in a supervised manner (see equation~\eqref{eq:end-to-end-training}). This requires pairs of measurement and target (ground-truth) images. However, in practice, such pairs cannot always be acquired, e.g., due to scan time constraints, signal decay effects along echo trains, or physiological motion. 
Therefore, self-supervised methods are attracting increased research interest. These methods make it possible to train networks without target or ground-truth data by either making assumptions on the measurements or using additional noisy or partial measurements. A plethora of approaches has been developed, including methods for learning from  under-sampled data \cite{Yaman2020,millard2022robustSSDU}, unpaired data \cite{chen2023deep}, or limited-resolution data \cite{wang2023k}. 
Here we describe several approaches that are architecture-agnostic. 
For recent reviews on this topic see \cite{akccakaya2022reviewSSL,zeng2021reviewSSL,chen2023deep}.

\subsubsection{Learning of algorithms based on Stein's Unbiased Risk Estimate (SURE)} \label{SURE_sec}
\label{denoiser}

We start with a method that is based on assumptions on the noise distribution, called Stein's Unbiased Risk Estimate (SURE)~\cite{sure}. We consider the estimation of $\mathbf x$, denoted by $\widehat{\mathbf x}$ from its noisy measurements $\vv = \mathbf x + \mathbf n$. Here $\mathbf n$ is zero-mean Gaussian noise with a variance of $\sigma^2$. In practice, the estimate $\widehat{\mathbf x}$ is derived from the noisy measurements $\vv$ using a deep network as $\widehat{\mathbf x} = f_{\phi}(\vv)$. When the noiseless reference image $\mathbf x$ is available, the true mean-square error (MSE), denoted by 
\begin{equation}\label{mse}
{\rm MSE} = \mathbb E_{\mathbf x} ~\|\widehat{\mathbf x} - \mathbf x\|^2
\end{equation} can be used.

By contrast, the SURE~\cite{sure} 
approach uses the loss function
\begin{equation}
\label{sure}
{\rm SURE}(f_{\phi}(\vv),\vv) = \| f_{\phi}(\vv)- \vv \|^2_2  + 2  \sigma^2 \nabla_{\vv} \cdot f_\Phi(\vv) - N\sigma^2, 
\end{equation}
which is an unbiased estimate of \eqref{mse}. 
Note that the expression in \eqref{sure} does not depend on the noise-free images $\mathbf x$; it only depends on the noisy images $\vv$ and the  network parameters $\phi$. In \eqref{sure}, $\nabla_{\mathbf u} \cdot f_\Phi(\boldsymbol{u})$ represents the network divergence, which is often estimated using Monte-Carlo simulations~\cite{mcsure}. Several researchers have adapted SURE as a loss function for the unsupervised training of deep image denoisers~\cite{ldampSURE,koreanReconCVPR2019} and demonstrated performance approaching that of supervised methods.

The SURE approach was extended to inverse problems with a rank-deficient measurement operator known as the generalized SURE~(GSURE)~\cite{eldarGSURE}. The GSURE provides an unbiased estimate of the projected MSE, which is the expected error of the projections in the range space of the measurement operator. The GSURE approach was recently used for inverse problems in \cite{ldampSURE}. The experiments in \cite{ldampSURE} showed that the GSURE-based projected MSE was a poor approximation of the actual MSE in the highly undersampled setting. To improve performance, the authors trained the denoisers at each iteration in a message-passing algorithm in a layer-by-layer fashion using classical SURE, which was termed LDAMP-SURE \cite{ldampSURE}. This approach approximates the residual aliasing errors at each iteration to be Gaussian random noise. As this assumption is violated in many inverse problems, the performance of this layer-by-layer training approach is not as good as supervised methods.

The ENSURE framework circumvents the poor approximation of the true MSE by GSURE by considering different sampling operators for different images. Similar to classical SURE metrics \cite{sureSink,eldarGSURE}, the ENSURE loss metric has a data consistency term and a divergence term. The data consistency term in ENSURE is the sum of the weighted projected losses \cite{eldarGSURE} from multiple subjects; the weighting depends on the class of sampling operators. When different sampling patterns from different subjects fully cover \emph{k-space}, the ENSURE metric is an unbiased estimate of the true image-domain MSE and hence is a superior loss function than projected SURE \cite{eldarGSURE}. The comparison of the above methods shows that the the ENSURE approach can provide performance comparable to that of supervised training.

%%%
\subsubsection{Self-supervised DL based on Noise2noise}

Noise2noise \cite{lehtinen_Noise2Noise} is a well-established framework, which  constructs a self-supervised loss based on independent noisy measurements of the same object. In the context of MRI, suppose we are given two undersampled measurements $\vy = \mM \mF \vx$ and $\vy' = \mM'\mF \vx$, where $\vx$ is the image of interest, $\mF$ is the Fourier transform, $\mM$ is a random undersampling mask and $\mM'$ is a different random mask, with each diagonal entry chosen independently. From these measurements, we can construct the self-supervised loss
\begin{align}
\ell_{\text{SS}}( f_\vth(\vy), \vy') 
=
\norm[2]{ \mM'\mF f_\vth(\vy) - \vy'}^2.
\end{align}
It can be shown that in expectation over the random measurements, a minimizer of the self-supervised loss is also a minimizer of the expectation of the supervised loss (see Prop. 2 in \cite{klug_AnalyzingSampleComplexity_2023_v2}). Thus, with enough training examples, such self-supervised training can approach the performance of supervised training~\cite{klug_AnalyzingSampleComplexity_2023_v2}. %There are different choices on how such a loss can be constructed. 

One notable method that has implemented this approach successfully for MRI reconstruction is  Self-Supervised Learning
via Data Undersampling (SSDU) \cite{Yaman2020}. This method partitions available \emph{k-space} measurements into two disjoint sets; the first set is used in the data consistency units of the unrolled network, i.e., for the forward pass, and the other one is used for computing the loss, i.e., for supervision. SSDU can hence be trained using under-sampled data alone. In their work, Yaman et al. \cite{Yaman2020} demonstrated that SSDU achieved comparable performance to fully supervised learning methods while offering practical advantages in real-world MRI applications. 

Recently, Millard and Chiew \cite{chiew2023theoretical} introduced a general theoretical framework that extends Noiser2Noise  \cite{moran2020noisier2noise} and also explains SSDU. Unlike the SSDU formulation, where one set is recovered from the other, they applied two subsampling masks to the data. They proposed a weighted $\ell_2$ loss, computed in \emph{k-space}, with a weighting that compensates for the sampling and sampling-partitioning densities. They derived the framework analytically and showed that when the weighting matrix $W$ is rank-deficient and fulfils certain conditions, the method boils down to SSDU. They showed analytically that SSDU with an $\ell_2$ \emph{k-space} loss approximates fully sampled reconstruction, on expectation. It is worth mentioning that their analysis was done for an $\ell_2$ \emph{k-space} loss, while the original SSDU method was trained with a mixed $\ell_1/\ell_2$ loss.

\subsubsection{Self-supervised DL using limited-resolution data (\emph{k-space} bands)}

The self-supervised methods described above focused on learning from under-sampled data acquired with variable-density or parallel-imaging schemes. Although such datasets have undersampling artifacts, they effectively constitute high-resolution data, because the sampling masks commonly cover the entire \emph{k-space} extent (note that under-sampling creates artifacts but does not necessarily reduce the resolution). However, the acquisition of high-resolution data can be challenging. In dynamic MRI, for example, there is often a trade-off between the spatial and temporal resolutions, which requires acquisition compromises. 

\begin{figure}[t]
\centerline{\includegraphics[width=0.5\textwidth]{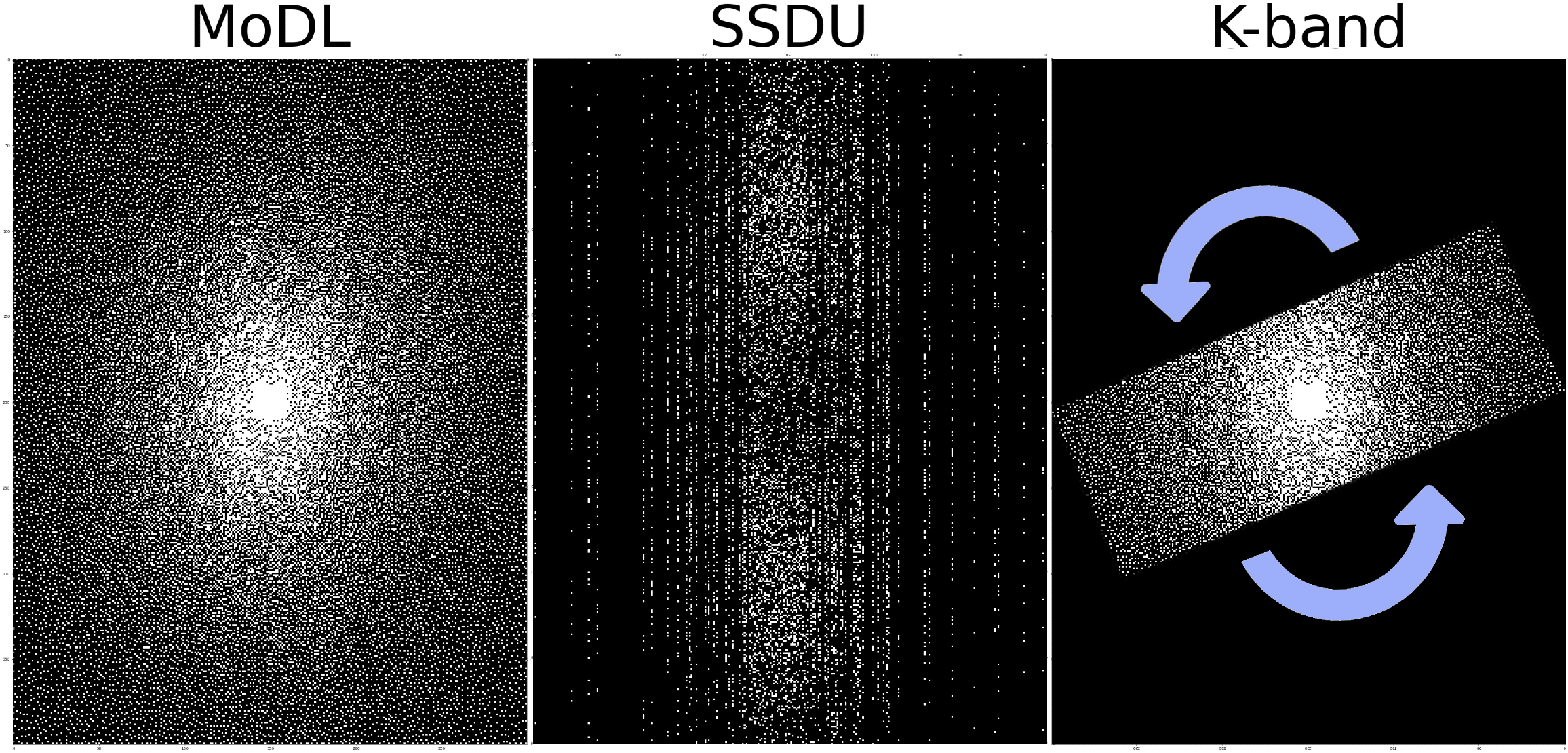}}
\captionsetup{width=0.8\textwidth}
%\centering
%\includegraphics{samp_comp.png}
    
\caption{Example of the input \emph{training} data for three DL reconstruction methods. The fully-supervised MoDL method \cite{aggarwal2018modl} receives var-dens sampled data as input and uses the entire \emph{k-space} for supervision.  The self-supervised SSDU method \cite{Yaman2020} receives var-dens data as input, splits it into two subsets, and uses one set for data consistency and the other for supervision. In this example, the var-dens data were sampled from parallel-imaging (equispaced) acquired data, as in \cite{Yaman2020}. The \emph{k-band} method \cite{wang2023k} receives var-dens sampled data from a \emph{k-space} band, and uses data from the whole band for supervision, without any supervision outside the band. Different bands are acquired from different subjects, with random orientations. At \emph{inference}, the input to all three methods is var-dens data from the entire \emph{k-space}, similar to that shown here for MoDL.}
\label{fig:samp_fig}
\end{figure}

Recently, the \emph{k-band} framework was proposed for self-supervised learning from partial, \emph{limited-resolution} data \cite{wang2023k,Wang2023kbandSedona}. This framework is based on the acquisition of \emph{k-space} bands, where each band acquires data with high resolution in the MRI readout dimension and limited resolution in the phase encoding (PE) dimension. The authors suggested acquiring different bands from different subjects, and randomizing the bands' orientation across subjects; fundamentally, this randomization serves to expose the network to all \emph{k-space} areas across the training iterations (Fig. \ref{fig:samp_fig}). Thus, even though the network does not get a full \emph{k-space} from any single subject, it can learn connections across all the \emph{k-space} regions. To enable self-supervised learning from limited-resolution data without limiting the resolution during inference, the authors introduced an optimization method dubbed \emph{stochastic gradient descent (SGD) over\emph{k-space}subsets}.

In this framework, the loss is computed in \emph{k-space} and formulated by
\begin{align}
\ell_{k\_band} = \|\mW \mB (\mF f_{\theta}(\vy) - \mF \vx ) \|_1
\end{align}
 where $\mB \in \{\mB_i\}_{i=1,...,180}$ is a binary band sampling operator that samples a band with angle $i$, and $\mW$ is the loss weighting mask
\begin{align}
%\label{eq:w_mask}
\mW = 180 (\sum_{i=1}^{180} \mB_i)^{-1}.
\end{align}
This loss-weighting compensates for the over-exposure of the network to low-frequency \emph{k-space} data and enhances learning in the \emph{k-space} periphery. This is beneficial because in the \emph{k-band} acquisition setting, the center of \emph{k-space} is included in all bands (Fig. \ref{fig:samp_fig}), unlike the periphery. The authors showed analytically that when this loss-weighting mask is applied, the self-supervised training process stochastically approximates fully supervised training, on expectation.  They demonstrated that learning from limited-resolution data can hence result in performance comparable to supervised and self-supervised methods trained on high-resolution data, and hence offers a practical solution for cases where such data are unavailable.

\subsubsection{Loss-weighting}

Both \cite{wang2023k} and \cite{millard2022robustSSDU} demonstrated that applying loss-weighting in \emph{k-space} can enhance the performance of self-supervised DL methods. Interestingly, these studies reached similar conclusions while analyzing different reconstruction frameworks. Millard and Chiew \cite{millard2022robustSSDU} analyzed SSDU with an $\ell_2$ \emph{k-space} loss, while
Wang et al. \cite{wang2023k} analyzed \emph{k-band} with an $\ell_1$  \emph{k-space} loss. Another difference is that Millard and Chiew  \cite{millard2022robustSSDU} used variable-density masks that cover the entire \emph{k-space}, while Wang et al. \cite{wang2023k} focused on \emph{k-space} band acquisition.
However, both studies derived loss-weighting masks that weigh down the loss in the center of \emph{k-space} and enhance it in the periphery. In other words, these masks inhibit the learning of low-frequency data and facilitate the learning of high-frequency details, so that eventually all frequencies are weighted equally. Given that different settings were analyzed, this strategy may be effective for other methods as well.

% =============== Advanced supervised approaches ===========

\subsection{Recent architectures: transformers and dual-domain networks}

In addition to the training methods described above, much progress has also been made in the development of advanced architectures. For example,
two architectures that recently garnered substantial attention are transformers and dual-domain networks. Transformers \cite{han2022survey,han2021transformer} have powerful computational capabilities due to their use of an attention mechanism \cite{vaswani2017attention} that makes it possible to weigh the importance of different parts of the input data and capture long-range dependencies. Transformers  first made a substantial impact in the field of natural language processing  \cite{gillioz2020overview,wolf2019huggingface} and then became highly influential in computer vision  \cite{han2022survey}.

In the context of MRI reconstruction, recent studies demonstrated that transformers offer excellent performance and ability to deliver improved structural and textural fidelity. For example, Korkmaz et al. \cite{Korkmaz2021} developed an unsupervised MRI reconstruction method based on a generative vision transformer. Their method utilizes cross-attention transformer blocks, which receive both global and local latent variables as input and progressively map them to MR images with increasing spatial resolution. This style-generative architecture enhances  representational learning and improves model invertibility. Feng et al. \cite{feng2021task} introduced the $T^2Net$ for simultaneous MRI reconstruction and super-resolution. This network has two branches dedicated to these two tasks, and incorporates a task transformer module to facilitate effective feature sharing between them. Guo et al. \cite{guo2023reconformer} introduced the ReconFormer, an architecture that leverages recurrent pyramid transformer layers and scale-wise attention mechanisms. It effectively captures multi-scale information and deep feature correlations, leading to efficient, high-quality image reconstruction and computational efficiency.

Another emerging type of architecture is known as dual-domain networks, which integrate information from the image and \emph{k-space} domains \cite{souza2020dual,eo2018kiki,wang2022DIMENSION,ran2020md,singh2022joint}. This approach, exemplified by MD-Recon-Net \cite{ran2020md}, leverages the complementary strengths of these two domains to enhance reconstruction quality. A study by Souza et al. \cite{souza2020dual} demonstrated the effectiveness of such networks in multi-channel MRI reconstructions.  Singh et al. \cite{singh2022joint} demonstrated that layers utilizing joint learning of image and frequency domain features can directly replace standard convolutional layers. This is useful for numerous tasks, including image reconstruction, motion correction and denoising.

Transformers and dual-domain networks have recently been integrated, leading to state-of-the-art architectures.  For example, Zhao et al. \cite{zhao2023swingan} introduced SwinGAN, a dual-domain Swin Transformer-based GAN. This network combines frequency-domain and image-domain generators, both utilizing Swin Transformer backbones. This design allows for effective capture of long-distance dependencies in MR images. SwinGAN also features a contextual image relative position encoder, which enhances its ability to capture local information. Wang et al. introduced DCT-Net, a dual-domain transformer network for MRI reconstruction \cite{wang2024dct}, which integrates image and frequency domain information through its cross-attention and fusion-attention blocks. DCT-Net is designed to enhance MRI reconstruction performance, particularly under low sampling rates, by leveraging the complementary strengths of both domains. In summary, these recent architectures offer high computational power to improve image reconstruction quality.

\section{DL for acquisition optimization}

In this section we review two applications in which DL is utilized to improve MRI acquisition. We first explore methods that optimize the \emph{k-space} sampling trajectories in tandem with the reconstruction, and then turn to advances in harnessing DL to refine  MRI pulse sequences. 

\subsection{Optimizing \emph{k-space} trajectories} %\blue{(Mathews, Efrat)}

The computational design of sampling patterns has a long history in MRI. Generally, two types of approaches have been taken.  \emph{Algorithm-agnostic} methods, e.g. \cite{Reeves2000,xu,haldar2019oedipus,levine2017,senel2019}, consider specific image properties (e.g., the Cramér-Rao bound or image support) and optimize the sampling pattern to improve the measurement diversity for that class. \emph{Algorithm-dependent} methods, on the other hand, e.g.  \cite{sherry2019,gozcu2018learning,sparkling2019MRM,ravula2023sampling}, optimize the sampling pattern assuming specific reconstruction algorithms. These are  typically CS algorithms, which employ regularizers such as TV, wavelet-domain sparsity, or pre-trained diffusion models \cite{ravula2023sampling}. 

The main challenge with the above computational approaches is their  high computational complexity. In particular, algorithm-dependent schemes need to solve the CS problem for each image in the dataset, to evaluate the loss for a specific sampling pattern. The design of sampling pattern thus involves a nested optimization strategy; the optimization of the sampling patterns is performed in an outer loop, while image recovery is performed in the inner loop to evaluate the cost associated with the sampling pattern.

DL provides an opportunity to speed up the computational design, because DL inference schemes enable rapid evaluation of the loss for each sampling pattern. This enables a joint strategy that simultaneously optimizes the acquisition scheme and the reconstruction algorithm. Early DL-based joint optimization schemes solved for a binary sampling mask  \cite{weiss2019pilot,bahadir2020deep}. The PILOT approach, for example, solved for the sampling density, assuming a sampling layer similar to variational auto-encoders \cite{weiss2019pilot}. The LOUPE method, on the other hand, learned the optimal sampling density in tandem with a reconstruction network \cite{bahadir2020deep}. It was first developed for 2D Cartesian imaging \cite{bahadir2020deep} and later extended to non-Cartesian sampling \cite{Chaithya2021sparklingEUSIPCO}. Other studies have focused on 3D Cartesian sampling with a variational reconstruction network \cite{zibetti2022alternating}.

More recent work represents the sampling locations $\phi$ as continuous variables and jointly solves for them and for the parameters of the DL algorithms. These methods consider a forward model $\mathbf A_{\phi}$, where $\phi$ denotes the sampling locations. This forward model may be 
represented either by using an analytical Fourier transform  \cite{aggarwal2020jmodl}  or a non-uniform Fourier transform \cite{Wang2022bjork}. We denote the reconstruction algorithm (which can be unrolled, direct inversion, or plug-and-play) by 
\begin{equation}
\hat{\mathbf x} = \mathcal M_{\theta,\phi}(\mathbf y),
\end{equation}
where $\theta$ denotes the parameters of the reconstruction algorithm and $\phi$ are the sampling locations corresponding to the forward model. Joint optimization schemes, e.g. \cite{aggarwal2020jmodl,Wang2022bjork}, are designed to optimize the sampling pattern $\Phi$ and the CNN parameters $\phi$ in tandem, i.e.,
\begin{equation}
\label{joint}
\{\theta^*,\phi^*\} =  \arg \min_{ \theta ,\phi} \sum_{i=1}^{N} \| \mathcal M_{\theta,\phi} \left(\mathbf A_{\phi}(\mathbf x_i)\right) -\mathbf x_i \|_2^2.
\end{equation}

Several methods have been developed within this framework. For example, J-MODL focuses on a model-based reconstruction and utilizes an unrolled network \cite{aggarwal2020jmodl}. In a different work, Wang et al. \cite{Wang2022bjork} parameterized trajectories with quadratic B-spline kernels, and performed optimization under penalties describing realistic MRI hardware constraints, e.g. the slew rate and gradient amplitude. This work was later extended to a generalized Stochastic optimization framework for 3D
NOn-Cartesian samPling trajectorY (SNOPY) \cite{wang2022snopy}, which can accomodate several optimization objectives. Chaithya and Ciuciu \cite{Radhakrishna2023} introduced the PROJeCTOR framework, which enables joint learning of  non-Cartesian trajectories and reconstruction networks by using a projected gradient descent algorithm. Alkan et al. \cite{alkan2020sampling} introduced joint sampling and reconstruction optimization through variations in information maximization, where they used an encoder to represent non-uniform sampling and a decoder in an unrolled neural network. Xie et al. \cite{puert2022sampling} introduced the PUERT method for learning probabilistic sampling patterns along with an interpretable reconstruction method; their learning module incorporated a dynamic gradient estimation strategy. Finally, Zou et al. \cite{zou2022jointDCE} demonstrated that joint optimization can reduce the bias and
uncertainty of pharmacokinetic parameter estimation in dynamic contrast enhanced MRI, and hence contribute to higher diagnostic value
. Altogether, these methods have shown significant benefits from jointly optimizing  the sampling pattern and the reconstruction algorithm.

\subsection{Pulse sequence design} 
The previous section focused on accelerating MRI scans via \emph{k-space} sub-sampling. The complementary element to this effort is the optimization of the remaining pulse sequence parameters by a set of radio-frequency (RF) powers, shapes, and duration that enable the shortest possible scan time, while retaining sufficient contrast, SNR, and consistency with conventional (and lengthy) alternatives. 

The pulse sequence design task was traditionally hand-crafted by MR experts, who combined  strong intuitions and an understanding of spin physics with mathematical solutions of the Bloch equations. While a remarkable number of contrast mechanisms and imaging schedules have been developed since the invention of MRI, the reliance on solvable differential equations severely limits our ability to reach a globally optimized schedule and reduce the scan time. Recent developments in DL architectures and computational frameworks have created new opportunities for the automatic and efficient optimization of rapid acquisition protocols. 

Zhou et al. \cite{zhu2018automated, zhu2019automated} introduced the representation of the Bloch equations as a computational graph. By treating each of the acquisition parameters as a neural network node weight, an efficient gradient-descent-based optimization was realized, where simulated signal trajectories were fed into the network, enabling an automatic generation of pulse sequences. The resulting protocols were characterized by non-intuitive gradient waveforms, where continuous off-resonant excitation  applied as the receive channel was continuously and simultaneously recorded. This approach yielded an ultra-short scan time for T$_1$/T$_2$ mapping at 1D. By expanding for 2D imaging, Lee et al. \cite{lee2019flexible} used automatic differentiation to optimize the Cramér-Rao Lower Bound (CRLB) of multiple-echo spin echo T$_2$ mapping, driven equilibrium single pulse observation of T$_1$ (DESPOT1) mapping, and the MRF IR-FISP sequence. 

\begin{figure}[t]
\centerline{\includegraphics[height=2.8in,width=5.5in]{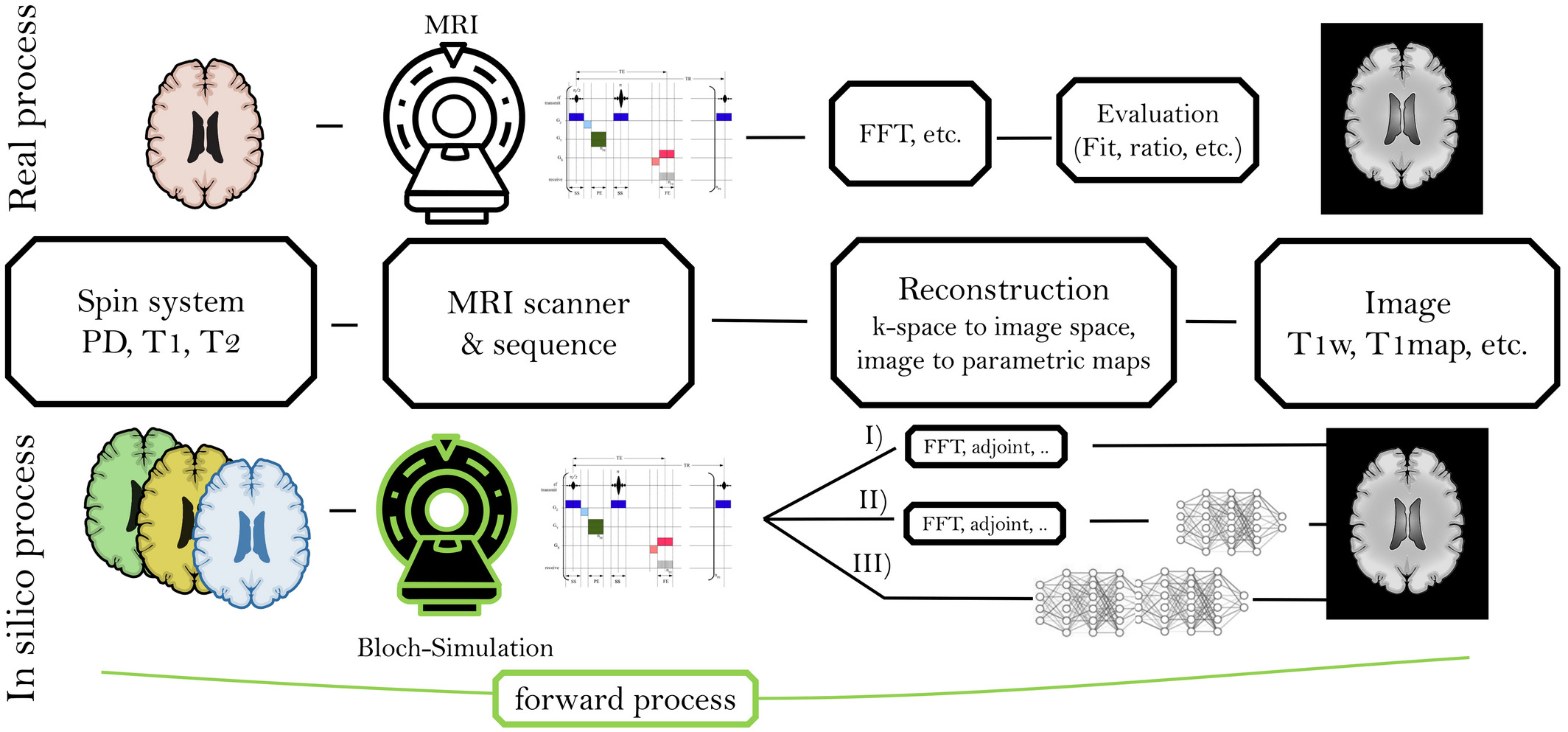}}
\captionsetup{width=0.85\textwidth}

\caption{\textbf{Automated discovery of MRI acquisition protocols using supervised learning}. A differentiable MR scanner utilizes the Bloch equations for in-silico signal generation and the later reconstruction of the target contrast of interest from real, acquired data. Reproduced from Loktyushin et al. \cite{loktyushin2021mrzero}.}
\label{fig:mrzero_fig}
\end{figure}

Loktyushin et al. \cite{loktyushin2021mrzero} developed a supervised learning framework termed MR-Zero where a target contrast of interest is used for learning the optimal set of RF events, the gradient moment, and the delay times (Fig. \ref{fig:mrzero_fig}). One important feature of this approach is the use of a task-driven cost function that provides the user with the flexibility to prioritize the characteristics required from the output protocol, such as high data fidelity, short scan time, or the specific absorption rate (SAR) limits. In a later study, the same group used this approach to optimize the refocusing flip angles and minimize T$_2$-induced blurring in accelerated spin echo sequences \cite{dang2023mr}.

In the molecular MRI field, an end-to-end DL-based framework was developed for the discovery of rapid, quantitative chemical exchange saturation transfer (CEST), semisolid MT acquisition and reconstruction protocols \cite{perlman2022end}. The system was based on a computational graph representation of the Bloch-McConnell analytical solution which receives the molecular imaging scenario of interest as input, and outputs an optimized set of acquisition parameters and the corresponding reconstruction network that  translates the raw data into quantitative parameter maps. In vivo experiments showed it could acquire in-vivo data in merely 35 seconds and reconstruct parameter maps in less than 1 second. The use of recurrent neural networks and training over a wide range of saturation pulse frequency offsets have further increased the robustness of this conceptual approach for B$_{0}$ and B$_{1}$ inhomogeneity \cite{kang2023only}. 

All these approaches exploit DL-based strategies to optimize and derive novel acquisition routes \textit{offline}. Recently, a different optimization paradigm was suggested where the acquisition parameters are modified and adapted on the fly, \textit{during} data acquisition \cite{beracha2023adaptive}. By combining a Bayesian framework, CRLB calculation, and model-based reconstruction, the acquisition parameters for a series of images can be optimized in real-time based on the previous image history. This concept has demonstrated up to a 3.3 fold acceleration of multi-echo sequences in human subjects and molecular imaging phantoms.

%%%
\section{Advanced techniques and applications}

In this section we discuss DL methods for quantitative MRI and  dynamic MRI. 

\subsection{DL methods for quantitative MRI} %} \blue{(Or Perlman)}}
The goal of quantitative MRI is to extract one or more tissue parameter maps from a series of qualitative images \cite{zhao2016maximum}:
\begin{equation}
\label{quat_task}
\mathbf I_m=\mathbf \Phi_m(T_{param})\mathbf \rho
\end{equation}
where I$_m$ denotes the contrast-weighted images for m=1,...,M acquisitions, $\rho$ denotes the spin density, T$_{param}$ denotes the tissue parameters (T$_1$, T$_2$, etc.), and $\Phi_m$ is the biophysical function connecting the acquisition parameters with the resulting contrast-weighted images.
The mapping of tissue properties enables de-biasing imaging protocols and harmonization of the final diagnosis across sites, vendors, and physicians. It thus provides sensitive and standardized tools for reproducible interpretation of MRI-based information \cite{seiberlich2020quantitative_v3}. 
 The classical approach to MR property mapping mandates repeated acquisition, where all the protocol parameters are held fixed, and only a single parameter is slowly and gradually varied (e.g., the flip angle or the repetition time across different M acquisitions). The resulting long scan time hinders the widespread use of quantitative MRI in clinical settings \cite{vladimirov2023molecular}. The reconstruction of the acquired raw data series demands a lengthy parameter-fitting procedure that is computationally intensive and slow.

The development of powerful DL architectures such as CNNs, UNets, GANs, ResNets and recurrent neural networks has been leveraged to accelerate and enhance the performance of quantitative MRI \cite{feng2022rapid}. To accelerate relaxometry studies, Liu et al. \cite{liu2019mantis} developed a model-augmented neural network that receives a series of incoherently-sampled multi-echo images and uses a CNN to reconstruct the T$_2$ parameter maps. The supervised learning was guided by a parameter-space loss, which compares the reconstructed T$_2$ maps to the ground truth reference, and a \emph{k-space} loss. The latter was designed to ensure that the physical-model-based synthetic undersampled \emph{k-space} measurements matched the originally acquired \emph{k-space} information. In a later work, the same group developed a model-guided self-supervised DL framework for rapid T$_1$/T$_2$ mapping \cite{liu2021magnetic}, to obviate the need for fully sampled training references.

While many DL-based quantitative mapping strategies are focused on \emph{k-space} sub-sampling, a further acceleration potential lies in reducing the number of contrast-weighted images acquired. In a very recent work, Li et al. \cite{li2023supermap} trained a deep residual CNN network to receive just three \emph{k-space} under-sampled contrast weighted images and output the corresponding T$_{1rho}$ and T$_2$ parametric maps (which are particularly useful for the study of osteoarthritis).

Magnetic resonance fingerprinting (MRF), which was first reported in a 2013 Nature paper \cite{ma2013magnetic} and increasingly studied since then, constitutes a paradigm shift in MRI-based tissue characterization. Unlike traditional relaxometry studies, MRF starts with the acquisition of tens or hundreds of images, using a pseud-random acquisition pattern accommodating a series of short repetition times, small flip angles, and heavily under-sampled \emph{k-space} data (e.g., via a single variable density spiral trajectory). Although each of the resulting raw images is extremely noisy, the temporal evolution of the signal at each pixel entails a unique fingerprint. By comparing the experimental trajectory to a Bloch-equation-derived dictionary of simulated signals, the inverse problem can be solved to uncover the associated parameter maps (T$_{param}$).

\begin{figure}[t]
\centerline{\includegraphics[width=0.8\textwidth]{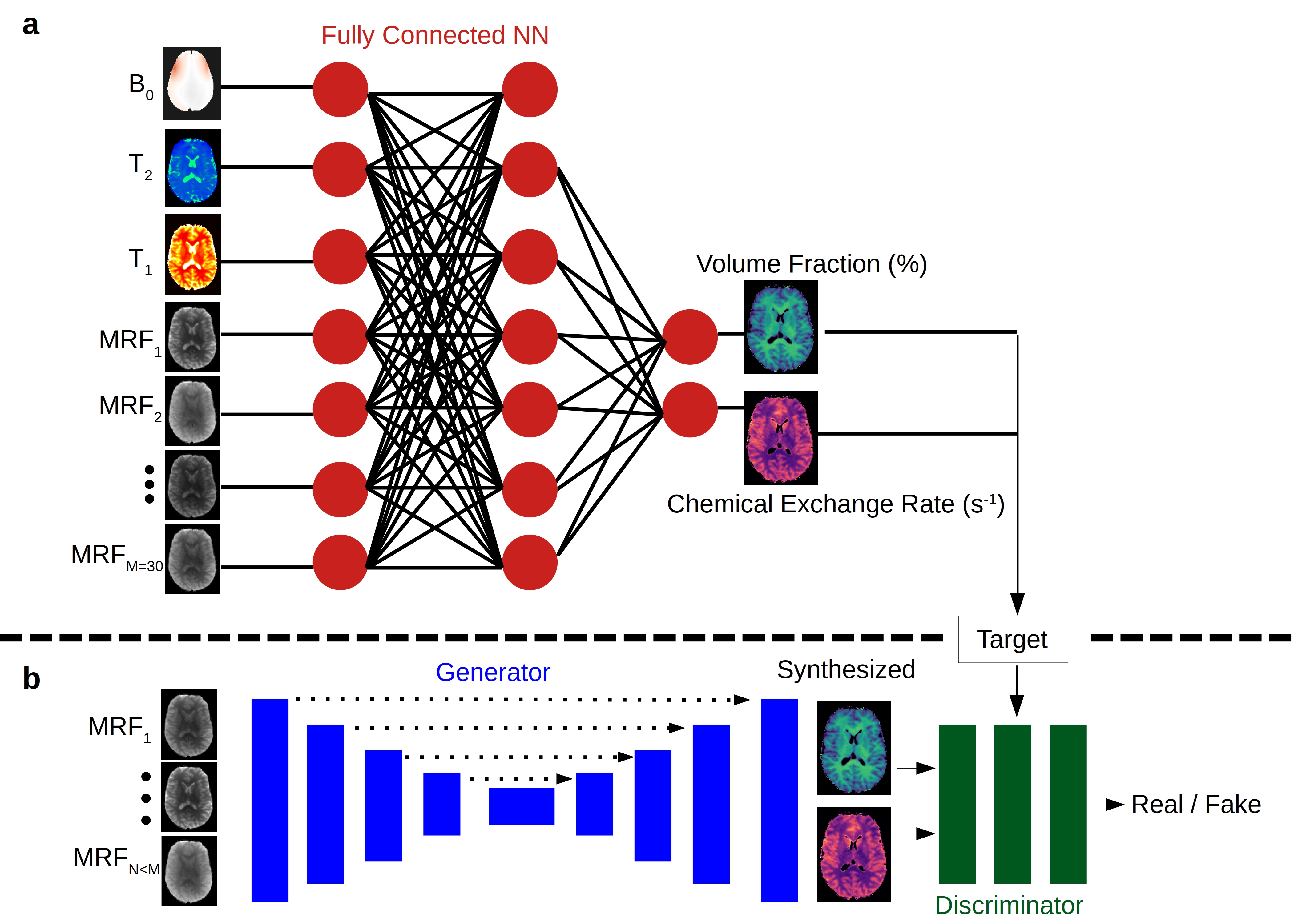}}
\captionsetup{width=0.85\textwidth}
\caption{\textbf{Deep learning reconstruction of quantitative magnetic resonance fingerprinting (MRF) information.} \textbf{(a)} A fully connected neural network is trained using simulated signal trajectories. During inference, it receives a series of raw MRF images pixel-wise, as well as auxiliary maps, yielding quantitative parameter maps. \textbf{(b)} A further acceleration in scan time can be achieved by training a generative adversarial network (GAN) using a smaller subset of raw input data to yield the same quantitative output maps. Reproduced and modified from Weigand-Whittier et al. %\textit{Magn. Reson. Med.} 2023; 89:1901-1914 
\cite{weigand2023accelerated}.}
\label{fig:qMRI_fig}
\end{figure}

While the resulting acquisition times are incredibly short (e.g., $<$13 s \cite{panda2017magnetic, ma2013magnetic}) the quantitative image reconstruction step (via pattern matching) may take hours, because the similarity between each acquired signal trajectory and \textit{all} possible dictionary entries needs to be calculated. 

In recent years, several DL-based strategies have been suggested to overcome this challenge. Cohen et al. \cite{cohen2018mr} trained a fully connected neural network using synthetic signal dictionaries to reconstruct MRF data in less than 100 ms. To take advantage of the inherent dependencies between adjacent image pixels, Balsiger et al. \cite{balsiger2018magnetic} designed a spatiotemporal CNN where the time-evolution dimension is the third dimension of the CNN patch kernel. In-vivo human brain validation studies using this approach demonstrated improved performance compared to alternative MRF networks \cite{cohen2018mr, gomez2016simultaneous}.
To accommodate 3D imaging, Gomez et al. \cite{gomez2020rapid} combined fully connected reconstruction networks with radial and spiral readout trajectories, and achieved whole-brain reconstruction in less than 7 min (compared to $>1.5$ hours using traditional reconstruction). 

Another MRF-associated bottleneck relates to the time required to generate the synthetic signal dictionary, which increases exponentially with the number of simulated parameters \cite{cohen2023cest}. Even when high-end computer clusters are used for this task, the computation time may reach hours/days for complex multi-pool imaging \cite{perlman2022quantitative, perlman2020cest}. Recently, this challenge was addressed by training a fully connected neural network using a variety of dictionaries to learn the nonlinear relations embedded in the physical model. The resulting system enabled the rapid generation of simulated signals for various protocols and imaging scenarios \cite{nagar2023dynamic}. NN-based simulators can be further combined with reconstruction networks to provide a unified rapid method for MRF analysis \cite{singh2023bloch}. A different approach to circumvent the need for exhaustive dictionary generation involves the direct synthesis of multi-contrast images (e.g., T$_1$-weighted, T$_2$-weighted, and FLAIR) from raw MRF data. While synthetic images can be derived from quantitative MRF data by forward model activation using the desired acquisition parameters \cite{blystad2012synthetic, perlman2023mr}, Wang et al. \cite{wang2023high} showed that a dictionary-free conditional GAN (trained on MRF raw data and paired ground truth weighted images) can perform the same task much faster. For cases where full quantitative information is required, a different work demonstrated that multi-parameter maps can still be extracted with GANs, even when merely 30$\%$ of the acquired MRF data is used \cite{weigand2023accelerated}  (Fig. \ref{fig:qMRI_fig}).

\subsection{Dynamic MRI}  %  \blue{(Efrat and Mathews)

Deep learning has become a transformative force in the realm of dynamic MRI, particularly in addressing the challenges related to limited acquisitions and motion correction, which constitute substantial hurdles in clinical imaging \cite{Zaitsev2015,Spieker2023,kustner2020CINENET,Qin2019CRNN,johnson2019conditional}. DL excels at learning signal evolution \cite{arefeen2023latent,wang2022DIMENSION}, a critical factor when aiming to accurately visualize and interpret dynamic changes in the body.

\subsubsection{Motion-resolved reconstruction}
Motion-resolved algorithms can effectively learn spatio-temporal correlations and reconstruct images from highly undersampled sequential data \cite{Biswas2019modl-SToRM,yoo2021time}. These methods have primarily been developed in the context of cardiac MRI \cite{kustner2020CINENET,Schlemper2018,poddar2019manifold,sandino2021deep,Biswas2019modl-SToRM,qin2018joint}. For instance, supervised unrolled algorithms have been used to recover cardiac cine MRI from breath-held MRI using using 4D (3D+Time spatial) convolutions \cite{kustner2020CINENET,Schlemper2018}. In other studies, architectures included unrolled algorithms that combine manifold \cite{poddar2019manifold} or low-rank priors \cite{sandino2021deep,Biswas2019modl-SToRM}, and joint learning of motion estimation and segmentation in cardiac MRI \cite{qin2018joint}. From a clinical perspective, DL has been found useful for measuring myocardial displacement \cite{ghadimi2021fully}, noninvasive diagnosis of myocardial ischemia \cite{scannell2020deep}, and  evaluation of cardiac function in pediatric imaging  \cite{zucker2021free}. 

To tackle the scarcity of training data, \emph{unsupervised} implicit learning approaches have recently been introduced for dynamic MRI \cite{yoo2021time,genStormTMI} (for more information see Section \ref{tdip}). These methods were also extended to multi-slice dynamic MRI data \cite{zou2022variational}, where the dynamic data from each slice $\mathbf x_i(t)$ are  acquired sequentially at different time points. The above model has been generalized to recover a pseudo-3D reconstruction by modeling the data as $\mathbf x_i(t)=\mathcal G_{\theta}\Big(\mathbf z_i(t)\Big)$, where $\mathbf z_i(t)$ are allowed to vary for different slices.

\subsubsection{Motion-compensated reconstruction}
 
The development of DL techniques for motion estimation and correction is a highly active research field, as DL can accurately detect and compensate for both rigid and non-rigid motion artifacts, which leads to more diagnostically valuable images. A thorough review of motion estimation and correction techniques is beyond the scope of this manuscript Here we highlight some of the main applications, and more information can be found in recent reviews \cite{sermesant2021natureCardiac,Spieker2023,Oscanoa2023,chang2023reviewDLrigid,zhao2021deep}. 

One of the main applications where DL is highly effective for motion correction is brain MRI, which is characterized by rigid-body motion \cite{chang2023reviewDLrigid,zhao2021deep,Haskell2019namer,Singh2023}. One of the early works in this field, by Johnson and Dragnova \cite{johnson2019conditional}, proposed conditional GANs to infer clean images from motion-corrupted data. More recent techniques include co-optimization for jointly estimating the motion parameters and reconstructed image \cite{Haskell2019namer,chen2023deep}, methods for detection and correction of motion-corrupted \emph{k-space} lines \cite{Eichhorn2023,Singh2023} and the use of score-based generative models \cite{levac2023accelerated}.

DL approaches are also highly useful for tackling \emph{non-rigid}, irregular motion. Applications include imaging of the trunk \cite{kustner2020deep_body_trunk}, fetal MRI \cite{uecker2021fetal_mri_review,singh2020deep}, abdominal MRI \cite{shimron2022bladenet,Murray2024movienet}, and MR angiography \cite{malave2020reconstruction}. Unsupervised implicit learning methods that represent the deformable motion fields at each time point have also been introduced and found to be effective in motion-compensated recovery \cite{mocostorm}. DL is also making strides in the field of real-time interventional MRI. Here, the rapid processing capabilities of DL algorithms enable real-time feedback and guidance during medical procedures, thus enhancing both the safety and efficacy of interventions. \cite{freedman2021dracula,terpstra2020deep,waddington2023real}

\subsection{Multi-task pipelines} %\blue{(Akshay)}

The fundamental goal motivating the acquisition of diagnostic-quality MR images is to extract clinically-used insights to further clinical care or to interrogate disease activity. Consequently, efficient, high-quality image acquisition is just the first step (typically referred to as \emph{upstream DL}) in the imaging workflow, which is followed by image analysis and insight extraction (typically referred to as \emph{downstream DL}) \cite{Sandino2021}. In many applications, these upstream and downstream processes are disconnected, leading to insufficient insights as to whether a novel MRI acceleration and reconstruction technique can reliably produce the requisite diagnostic information \cite{Chaudhari2020}. As a result, there is a substantial need to combine the upstream and downstream processes to ultimately harness advances in MRI physics, hardware, and DL for end-to-end acquisition-to-analysis workflows. 

Conventional and DL-based reconstruction techniques can potentially be combined with downstream task of clinical utility to guide useful model development. Specifically, MRI reconstruction workflows can be combined with three different downstream tasks that use whole images as inputs: (i) \emph{image classification}, which performs binary identification (via a yes/no) to identify the presence of one or multiple disorders; (ii) \emph{abnormality detection}, which performs localization via bounding boxes to accurately depict where one or multiple disorders are present in images; (iii) \emph{image segmentation}, which performs image classification at the voxel level to distinguish voxels that belong to a particular tissue or disease class. The DL sub-field of multi-task learning can learn multiple tasks simultaneously with \emph{positive task transfer}, where learning one task improves the performance of other tasks. 

In the context of accelerated MRI, combining the upstream task of MRI reconstruction with the downstream tasks of classification, detection, or segmentation can improve performance on all tasks. It also contributes to optimizing reconstruction techniques with clinically informed metrics. One of the greatest challenges in doing so, however, is the lack of available datasets that can merge both sets of tasks. The fastMRI raw-data dataset was recently supplemented with the fastMRI+ dataset that includes classification and detection bounding box annotations for knee and brain abnormalities at the slice level \cite{Zhao2022}. Such datasets can enable the design of end-to-end techniques to optimize reconstruction, subject to high performance on lesion detection \cite{zhao2021endtoend}. Similarly, even beyond end-to-end methods, such abnormality labels can be used to design clinical task-specific undersampling trajectories \cite{Weber_2024_WACV}. 

Beyond fastMRI+, SKM-TEA datasets include raw \emph{k-space} data as well as classification labels, detection bounding boxes, segmentation masks, and quantitative T2 relaxation time maps \cite{desai2021skm}. The original work profiled how different reconstruction approaches combined with different segmentation tools affected a common musculoskeletal biomarker of cartilage T2 relaxation time. Despite differences in the performance of the individual DL blocks, the overall impact on regional cartilage T2 values was small, a surprising finding that has been replicated for cartilage morphology and T2 tasks \cite{Desai2021seg, Schmidt2022}. Recent work has evaluated new approaches that combine generic pre-training tasks such as image reconstruction with fine-tuning for different clinically-relevant downstream tasks \cite{wu2023learning}. This approach achieves high performance in image acceleration as well as segmentation. Similar to these findings, the K2S challenge at MICCAI 2022 combined knee MRI reconstruction with bone/cartilage segmentation and bone shape analysis \cite{tolpadi2023k2s}. Yet again, there were only weak correlations between the metrics of reconstruction and segmentation quality, with one of the best segmentation models producing highly artifactual reconstructions but high quality segmentations.

\subsection{Joint estimation of sensitivity maps and reconstruction} % \blue{(Efrat)

The power of DL has also been harnessed for improving parallel multi-coil MRI, where the coil sensitivity maps must be estimated and incorporated in the image reconstruction process. Early DL methods commonly utilized the ESPIRIT algorithm \cite{uecker2014espirit} for computing the sensitivity maps prior to the reconstruction process. Nevertheless, joint estimation of the sensitivity maps and reconstructed data could contribute to improving image quality, as indicated in different studies, first with classical approaches\cite{uecker2008image,ying2007joint} and later using DL  \cite{sriram2020end,jun2021joint,luo2021joint}. DL frameworks were hence recently developed for joint estimation of the sensitivity maps and reconstruction data. For example, the well-known E2E-VarNet method \cite{sriram2020end} included a module for sensitivity maps estimation and incorporated it into a larger unrolled network, trained end-to-end. A similar approach was taken by Jun et al., who proposed the IC-Net \cite{jun2021joint}. Luo et al. suggested  using a deep image prior \cite{luo2021joint}, and Istanet et al. proposed a zero-shot learning method which is trained solely on data from a specific subject and jointly estimates the sensitivity maps and temporal data \cite{elfar2023zero}. %Furthermore, Zach et al. \cite{zach2023stable} 

\subsection{Other applications}

The powerful capabilities of DL have also been exploited for other computational tasks in the MRI workflow aside from image reconstruction. A detailed review of these applications is beyond the scope of this manuscript, which focuses on MRI reconstruction. Specific examples include the joint recovery of multi-contrast MRI data \cite{liu2023one,levac2023mri}, the synthesis of missing contrasts or synthesis of quantitative maps based on anatomical data \cite{liu2023one,sveinsson2021synthesizing}, super-resolution \cite{zhao2019applications,li2021review}, B0 estimation and off-resonance correction \cite{haskell2023off}), enhancement of low-field MRI data, where the low SNR degrades image quality \cite{ayde2022deep,koonjoo2021boosting}, and automated scan prescription \cite{lei2023automated}.

\section{Datasets and software} % \blue{(Efrat)}}

\subsection{Datasets}

The availability of public datasets and open-source code repositories has played a crucial role in the rapid development of DL techniques \cite{Bell2023}. In the MRI reconstruction field, several major databases such as  fastMRI \cite{knoll2020fastmri}, SKM-TEA \cite{desai2021skm}, mridata.org \cite{ong2018mridata}, and Calgary-Campinas \cite{souza2018open}  catalyzed development by making available large amounts of raw \emph{k-space} data, which are useful for developing and benchmarking methods \cite{ramzi_BenchmarkingDeepNets_2020_v2,hammernik2021systematic}. Other resources provide valuable data for specific applications, including MR imaging of speech production \cite{lim2021multispeaker}, cardiovascular imaging \cite{OCMR}, and low-field MRI \cite{wu2023m4raw_v2}. Many other MRI datasets  are also available on the web, but those were commonly designed for downstream, non-reconstruction tasks, hence they do not always contain raw \emph{k-space} data. Examples include the Human Connectome project 
\cite{human_connectome@2021}, IXI \cite{IXI}, BRaTS \cite{menze2014multimodal, ghaffari2019automated}, ADNI \cite{adni2008}, UK-biobank \cite{ollier2005ukbiobank} and OASIS \cite{oasis2017} 
%and others \cite{heath1998current,heath1998current,mias2016}. 

\subsection{Open-source software} %\blue{(Efrat)}}

The adoption of open-source frameworks has significantly accelerated the development of DL methods as they proide researchers robust, flexible platforms to develop new algorithms. The two most prominent general-purpose DL frameworks are PyTorch \cite{paszke2019pytorch} and TensorFlow \cite{tensorflow2015-whitepaper}, which offer extensive libraries that facilitate the design, training, and deployment of DL models. 

Several open-source software frameworks have been developed specifically for MRI. These offer useful computational tools for handling raw \emph{k-space} data, implementation of algorithms and computation of MRI-related metrics. For example, BART (Berkeley Advanced Reconstruction Toolbox) \cite{uecker2015berkeley} is a large and highly popular software package. It enables efficient data processing and contains implementations of different iterative reconstruction algorithms. The recent versions of BART also contain general-purpose tools that are highly useful for the development of DL reconstruction models, e.g., an automated differentiation framework compatible with complex-valued data, and implementations of well-established DL models \cite{blumenthal2023deep}. Gadgetron \cite{hansen2013gadgetron} is another popular package, which offers extensive tools for image reconstruction, data management, and implementations of iterative solvers. Another example is Sigpy \cite{ong2019sigpy}, which offers a set of operators, blocks and algorithms that are highly suitable for iterative reconstruction. Unlike other toolboxes, Sigpy is written entirely in Python and can hence be integrated easily into frameworks such as PyTorch.
The SNOPY (non-Cartesian sampling trajectory) \cite{wang2022snopy} framework has practical tools for  optimizing \emph{k-space} sampling trajectories, including a differentiable MRI system model, and loss functions corresponding to constraints on image quality, hardware (e.g., maximum slew rate and gradient strength), and peripheral nerve stimulation (PNS).
A different framework is Yarra (\url{https://cai2r.net/resources/yarra/}), which provides tools for automated collection of raw \emph{k-space} data. These can facilitate acquisition of datasets and the creation of new datasets.

Another key area is pulse sequence development. One of the main challenges in reproducing complex pulse sequences across different sites and scanners is the dedicated prototyping environment and software used by each vendor. Pulseq \cite{layton2017pulseq} is a rapid, hardware-independent pulse sequence prototyping framework, which   enables intuitive high-level programming of acquisition protocols in Matlab or Python. It enables easy deployment  across different field strengths and hardware. Importantly, the spin physics associated with the specific acquisition protocol compiled at the scanner can be accurately simulated as part of the Pulseq framework or its derivatives \cite{herz2021pulseq}. Techniques for data harmonization can help mitigate challenges in transferring protocols across different systems \cite{ning2020cross}. 

In the context of quantitative imaging, the qMRLab software was developed to facilitate reproducibility across MRI systems \cite{karakuzu2020qmrlab}. It consists of practical tools for analyzing and processing quantitative MRI data acquired by different vendors. The user-friendly interface and modular design of qMRLab enable researchers to easily implement and share quantitative MRI techniques.

Many other open-source codes can also be found on online platforms such as GitHub, articles with code (\url{https://paperswithcode.com/}), and the two dedicated websites of the ISMRM: MR-Hub  (\url{https://ismrm.github.io/mrhub/}), which hosts toolboxes, and MR-Pub (\url{https://ismrm.github.io/mrpub/}), which hosts git repositories published together with articles. 

The toolboxes and platforms described above are essential for enhancing reproducible research in the MRI community. The development of a unified data format, the ISMRMRD \cite{inati2017ismrm} can also facilitate easy translation of datasets and methods across sites and research groups. The recent development of techniques for federated learning \cite{elmas2022federated,levac2023federated} are also useful for training algorithms collaboratively without sharing the data; this can help address data-privacy issues.

% ============== Robusntess section ============
\section{Robustness challenges}

In this section we discuss challenges related to developing, evaluating, and benchmarking DL reconstruction methods, and suggest approaches for mitigating these challenges.

\subsection{Distribution shifts} % \blue{(Reinhard)

In deep learning, \emph{generalization} refers to the ability of a trained model to accurately reconstruct images that it has never seen before, particularly when these new images differ substantially from the training data. Good generalization of DL-based MRI reconstruction models is critical for clinical workflows. However, achieving a good generalization is challenging because MRI data can vary substantially in terms of different factors, e.g. MRI hardware, vendor-specific scanning protocols, patient populations, and the anatomical regions being imaged. This variability can lead to a model that performs well on data from one source but poorly on data from another, a phenomenon known as \emph{domain shift} or \emph{distribution shift}. Here we review  some of the main challenges related to this issue. 

Domain shifts have been studied from several perspectives.  \cite{johnson2021evaluation} analyzed the models submitted to the 2019 fastMRI challenge and found that many of them were sensitive to distribution shifts.  Darestani et al. \cite{darestani2021measuring} evaluated the robustness of DL reconstruction methods with regard to out-of-distribution data, and found that both trained and untrained networks were affected by distribution shifts. Avidan et al. studied another type of distribution shift \cite{avidan2022physically},  related to sampling; methods trained on specific sampling schemes may not generalize well to other schemes. Altogether, distribution shifts can lead to substantial performance drops in MRI and can hence be a major limiting factor in practice. 

\textbf{\emph{Potential mitigation strategies.}} In  cases where only a few training examples from a target domain are available, pre-training a network on other data and fine-tuning it to the target domain can improve performance~\cite{knoll2019assessment, huang2022evaluation,dar2020transfer}. 
In the challenging case where no target data are available for fine-tuning, test-time-training, which involves adapting to a single training at inference, is a viable performance-enhancing alternative~\cite{lin2023robustness}. Another good strategy that can help mitigate the performance drop due to distribution shifts is to train on broad and diverse data~\cite{lin2023robustness}. 

A growing body of work has pointed out the advantages of diffusion models in robustifying networks to distribution shifts.
As described above (section \ref{diffusion_models_section}), diffusion models decouple the image prior from the statistical measurement model. They can hence  generalize easily to various anatomies and sampling patterns \cite{jalal_RobustCompressedSensing_2021_v2,zach2023stable,gungor2023adaptive,chung2022score,daras_SolvingInverseProblems_2021_v2,yu2023universal}. Another technique to robustify networks to shifts in sampling patterns is to provide the network with the undersampling mask, and train the network to generalize to various sampling masks \cite{avidan2022physically}. In addition,  generative networks can be trained solely on magnitude images and applied to complex-valued data with  different sub-sampling patterns and out-of-distribution data \cite{zach2023stable}.

\subsection{Bias and "data crimes"} % \blue{(Efrat)

%Limited data → Bias due to off-label data use 
%“Off label” data use → data crimes

%Improper database construction - gender/ethnicity bias 
In the field of AI, the term \emph{bias} is often associated with gender-related or population-related bias. This can occur when models are trained on datasets that do not contain equal distributions of subjects having different genders, different ethnicities, or even  data that only contain a narrow set of medical conditions \cite{varoquaux2022machine}. This training can lead to algorithmic failure for under-served populations  \cite{seyyed2021underdiagnosis, chen2023algorithmic} or rare conditions \cite{varoquaux2022machine}.

However, when solving inverse problems, bias can also arise from a naive, seemingly-appropriate use of open-access datasets. One of the primary challenges in the development of DL reconstruction methods is the need for raw \emph{k-space} data, which are scarce and difficult to acquire due to the high cost of MRI scans and the long scan duration. While several databases offer such data, e.g. \cite{knoll2020fastmri,ong2018mridata,souza2018open,desai2021skm,lim2021multispeaker,wu2023learning}, there are many other databases that offer non-raw MRI data. Those are generally designed for downstream tasks, e.g., segmentation and classification, hence they are frequently preprocessed. Nevertheless, due to their high availability, researchers sometimes download and use them for synthesizing \emph{k-space} data training DL reconstruction models. This has been referred to as "off-label" data use, because those datasets are used for a different task than the one they were  were designed for \cite{shimron2022implicit}.

Surprisingly,  training DL reconstruction algorithms using "off-label" data could give rise to good-looking results; nevertheless, those are often biased and \emph{overly optimistic}, i.e. they are "too good to be true"  \cite{shimron2022implicit}. This is due to subtle preprocessing steps which, although imperceptible to the naked eye, impact algorithm performance. Common preprocessing steps include \emph{k-space} zero-padding, coil combination, and JPEG data compression. The authors of \cite{shimron2022implicit} demonstrated that  CS, dictionary learning, and DL algorithms are all sensitive to these preprocessing steps and yield biased results. This underscores their potential risk, as these algorithms are aimed for clinical purposes. The findings also demonstrated that DL algorithms trained on preprocessed data may not be able to generalize well to real-world clinical data, and could potentially eliminate crucial clinical details \cite{shimron2022implicit}. 

Another concerning finding is that 
popular error metrics, e.g., the normalized root mean square error (NRMSE) and SSIM could be \emph{blind} to the preprocessing \cite{shimron2022implicit}, and miss the bias. This is because those metrics compare the reconstructed and reference image, but those come from the same underlying data, i.e. both are preprocessed. The error metrics thus cannot measure the true image quality. Therefore, DL algorithms may achieve "good" NRMSE and SSIM scores even when their performance is poor. This makes the head-to-head comparison of results across papers very difficult, because some papers report experiments with raw \emph{k-space} data while others report results for preprocessed data, where the metrics tend to be better. To raise awareness of this phenomenon, the authors dubbed the publication of misleading results "data crimes".

\textbf{\emph{Potential mitigation strategies.}} In the context of gender-related or population-related biases, the best strategy is to train on large, diverse datasets \cite{seyyed2021underdiagnosis, chen2023algorithmic,varoquaux2022machine}.
In the context of bias that stems from training on preprocessed data, the optimal strategy is to train solely on raw \emph{k-space} data. When raw, zero-padded \emph{k-space} data are available, and such data have not been subject to any other preprocessing steps, one simple correction step is to crop \emph{k-space} to its original size. However, this scenario is quite rare. In practice, other preprocessing steps are often applied, e.g. coil-combination (e.g. using root-sum-of-squares operation) and JPEG compression. Those steps are irreversible, hence there is no simple technique to remedy them, and the bias cannot be easily prevented. 

\textbf{\emph{Techniques for data synthesis}.}
When raw \emph{k-space} data are unavailable, one possible strategy is to train on synthetic data \cite{deveshwar2023synthesizing,yang2022model,luo2023generative}. When magnitude data are available, synthetic complex-valued data can be obtained, for example, by adding a synthetic phase to the magnitude images and training a generative model to learn priors of complex-valued images \cite{luo2023generative}. Multi-coil data can be synthesized by multiplying the phase-enhanced magnitude data with sensitivity maps \cite{deveshwar2023synthesizing}. Another approach is to leverage the Bloch equations to simulate realistic data \cite{yang2022model,zhang2023towards}. However, it should be emphasized that the synthesis of complex-valued data does not guarantee good performance for real-world data; i.e., it cannot automatically prevent the bias described above. Models trained on synthetic data must therefore be tested using real-world, raw \emph{k-space} data.

\subsection{Hallucinations} %  \blue{(Efrat)}}

The term \emph{hallucinations} refers to the generation of false, realistic-looking features which are not present in the actual data. This can arise from the use of inaccurate priors \cite{bhadra2021hallucinations}, e.g., when there is a distribution shift between the training and test data, as described above. Strikingly, the team that organized the second FastMRI challenge found that many of the top-performing models produced hallucinations, and that these hallucinations were not captured by image quality metrics such as SSIM \cite{muckley2021results}. They also noticed that hallucinations could morph abnormal structures into seemingly normal ones. 

Several studies have also highlighted and explored manifestations of hallucinations. For example, Cohen et al. (2018) \cite{cohen2018distribution} discussed how distribution matching losses in medical image translation can lead to hallucinated features.   Bhadra et al. \cite{bhadra2021hallucinations} reported hallucinations in the context of tomographic image reconstruction and introduced the concept of hallucination maps to identify and understand the impact of prior information in regularized reconstruction methods. Gottschling et al. \cite{gottschling2020troublesome} explored hallucinations from a theoretical perspective, and highlighted problematic scenarios of \emph{in-distribution} hallucinations. The issue of hallucinations is a critical concern in DL-based image reconstruction, as it can lead to misleading results and false diagnoses. 

\textbf{\emph{Potential mitigation strategies.}} At present, there is a pressing need for techniques to mitigate hallucinations. Currently, the best strategy is to have radiology experts evaluate the reconstructed images, in the hope that they will be able to detect hallucinations. However, more research and development are required.

\subsection{Adversarial attacks and instabilities} % 
Neural networks for image classification are known to be sensitive to small, inperceptible, adversarially chosen perturbations. This has spurred concerns that DL-based reconstructions could be sensitive to worst-case perturbations. A number of studies~\cite{cohen2018distribution,huang2018some,gottschling2020troublesome,antun2020instabilities} have raised concerns regarding this issue. For example,  Antun et al. \cite{antun2020instabilities} provided simulations demonstrating that small, adversarially selected perturbations in undersampled measurements can result in severe reconstruction artifacts. 
They also showed that there is a trade-off between robustness and performance,  and that classical sparsity-based reconstruction methods are also sensitive to adversarially selected perturbations. 

Robustness to adversarial attacks is still an open issue, which requires further investigation. For example, at present there is no conclusive evidence indicating that DL reconstruction methods are more sensitive than classical methods such as sparsity-based reconstruction to worst-case perturbations.
\cite{krainovic_LearningProvablyRobust_2023_v2} provided theoretical results on a worst-case optimal estimator.  \cite{morshuis2022adversarial} found that simple end-to-end variational networks are as sensitive to perturbations as U-nets, and \cite{darestani2021measuring} showed that both un-trained and trained networks are sensitive to such perturbations. 

\textbf{\emph{Strategies for enhancing robustness.}}
Goujon et al. \cite{goujon2023neural} demonstrated that the robustness of reconstruction algorithms can be improved by constraining the CNN module to be convex or using a monotone constraint \cite{mol}. However, a global convexity or monotone constraint often translates into reduced performance \cite{goujon2023neural,mol}, as predicted by \cite{antun2020instabilities}. A recent work showed that the global monotone constraint described in \cite{mol} can be replaced by a local constraint around the image manifold \cite{john2023local}, to achieve improved robustness without compromising on performance. Following the approach implemented in non-convex algorithms, this algorithm is theoretically guaranteed to converge to a minimum, provided it is initialized with SENSE reconstructions. From a different perspective, \cite{caliva2020adversarial} and \cite{Cheng2020} suggested the use of adversarial attacks during training to reduce false negatives. 

% Benchmarking
\subsection{Benchmarking challenges} 

When evaluating various techniques  comparing the performance of different reconstruction and image analysis methods, it is crucial to have appropriate benchmarks that can characterize true clinical utility. Conventionally, image quality metrics such as the mean square error (MSE), peak signal-to-noise ratio (PSNR), and the structural similarity metric (SSIM) \cite{wang2004image} are used to assess the image quality of reconstructed MR images. These metrics have been described extensively in the computer vision literature and have a moderate correspondence with human-perceived image quality \cite{wang2004image}. However, as there is a substantial domain shift between natural images and medical images,studies have shown that such traditional image quality metrics do not correlate well with radiologist-perceived metrics \cite{mason2019comparison}. One likely reason for this phenomenon is that not all pixels in a given image have similar diagnostic value. Consider, for example, knee MRI scans, e.g. those contained in the popular fastMRI database.  Most abnormalities in knee scans are likely to be located in small, subtle regions in the cartilage, meniscus, and ligaments, while the tissues that are visible in the scans,  such as bone and muscle, are likely to have substantially fewer abnormalities. Computing traditional image quality metrics that weigh all image pixels similarly may therefore not reflect how a radiologist perceives the images. Thus, it is essential to define metrics that correspond to downstream clinical utility so that the same metrics can be used for benchmarking models. 

Another issue of concern is that the performance of DL reconstruction algorithms is typically evaluated using a relatively narrow test set, which is similar in nature to the training data. The evaluation results can thus be misleading, since they do not yield a reliable estimation of the model's generalization ability; i.e., the ability to perform well on test data that deviate from the training data, which is of critical importance in clinical settings. 

\begin{figure}[tb]
\centerline{\includegraphics[width=0.8\textwidth]{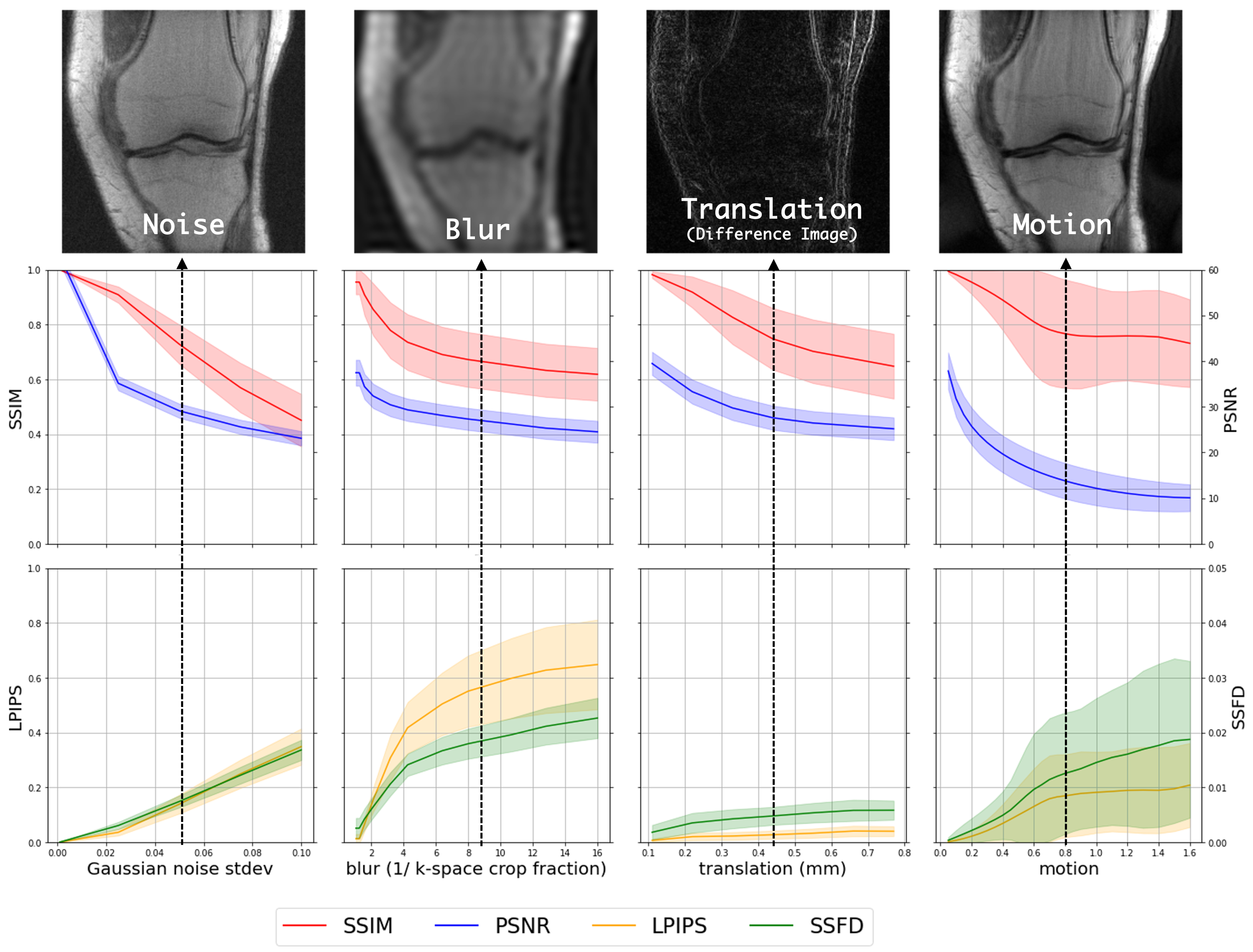}}
\captionsetup{width=0.85\textwidth}
\caption{\textbf{Impact of common image perturbations on image quality metrics}. A variety of image perturbations applied to a sample image from the fastMRI dataset (top row: noise addition, image blurring, pixel \textit{rolling} (where an image is shifted by a number of pixels), and physics-based subject motion. The impact of these corruptions is shown for conventional image quality metrics (SSIM, PSNR) and deep feature distance metrics (LPIPS - made for natural images, SSFD - made for MR images). The deep feature metrics exhibit a larger dynamic range to the noise, blurring, and motion corruptions, but present very little change due to pixel rolling, since the image quality does not change. These qualities of deep feature metrics are ideal for assessing MRI reconstruction quality.}
\label{fig:img_metrics}
\end{figure}

Yet another challenge is related to the common train/test data split. Even for a single dataset, different research groups or studies may apply different splits, and hence train and test the model with different data subsets. For example, even if two identical studies  used the same number of training images, it is not always possible to verify whether they were the same images. Previous work has shown that there can be a significant variation in how well the same deep learning models perform on different imaging exams within the same larger dataset \cite{darestani2021measuring}. Consequently, even if the number of training examples is maintained across studies, some examples may be easier or more challenging to train and evaluate. This makes a head-to-head comparison of published literature difficult to interpret. 

\emph{\textbf{Strategies for mitigating benchmarking challenges.}}
Initiatives such as the large-scale fastMRI database and challenge \cite{knoll2020fastmri} allow multiple different methods to be compared on identical training, validation, and testing data splits. This type of consistent evaluation platform allows for head-to-head comparisons across different methods and can shed light on their pros and cons.

More recent studies dealing with image quality assessment have gone beyond PSNR and SSIM metrics and used perceptual metrics of image quality instead of relying on handcrafted metrics \cite{johnson2016perceptual}. Perceptual metrics utilize representations extracted from pre-trained neural networks and have been used for evaluating and optimizing MR reconstruction quality \cite{adamson2023using, Wang2022}. For example, Learned Perceptual Image Patch Similarity (LPIPS) is commonly used for assessing the quality of natural images. A similar analog of Self-Supervised Feature Distance (SSFD) has been proposed for assessing MR image quality \cite{zhang_UnreasonableEffectivenessDeep_2018_v2,adamson2021ssfd_v2}. When using these perceptual metrics, a reference image and an evaluation image can be fed into the same pre-trained network, and the differences in their representations compared to compute the similarity between the two images. These perceptual metrics can either be computed using pre-trained networks on natural images or medical images. A recent study demonstrated that although there may not be considerable differences between networks trained on natural images and medical images; i.e., between the networks used for generating the representations, using perceptual metrics is better than using traditional image quality metrics when evaluating against radiologists' perceived image quality \cite{adamson2023using}. The impact of common image perturbations on such deep feature metrics is shown in Fig. \ref{fig:img_metrics}.

Beyond the metrics of image quality, another approach for benchmarking reconstruction models aims to directly benchmark the downstream value provided by the MR image. For example, the SKM-TEA dataset is designed to enable the evaluation of  downstream tasks such as clinical classification, segmentation, and articular cartilage T2 quantification  \cite{desai2021skm}. This dataset emphasizes the combination of automated cartilage segmentation with T2 quantification,  a known metric with clinical and research significance \cite{Chaudhari2019Rapid, chaudhari2021diagnostic}. Similarly, the K2S dataset aims to optimize cartilage volume and thickness quantification, whereas the fastMRI+ dataset assesses classification and detection tasks \cite{tolpadi2023k2s, zhao2021endtoend}. Such benchmarking metrics, which relate to clinical significance, provide a promising path forward for assessing image quality when used in conjunction with traditional or perceptual image quality metrics. Doing so can contribute to optimizing both image quality and image value for  radiologists.

\section{Uncertainty Estimation}

Connecting both upstream and downstream DL tools with clinical practice first requires promoting confidence among the users of these tools. Robustly characterizing the uncertainty of DL models may help increase the trust of downstream users in the outputs of DL algorithms prior to their integration into routine clinical practice. Uncertainty quantification, a popular subfield of DL research, can encourage building such trust among downstream users, regarding both image reconstruction and automated image analysis algorithms.

Several works have proposed methods to evaluate the uncertainty of accelerated MRI reconstructions \cite{tezcan2022sampling, edupuganti2020uncertainty, narnhofer2021bayesian,kustner2024predictive,luo2023bayesian,wangrigorous}. Given that MRI construction is an ill-posed problem where several high-quality images can correspond to the same low-quality image, quantifying the uncertainty of a sample reconstruction can help guide the fidelity of the overall reconstruction process. The developed uncertainty approaches  leverage the variational formulation of inverse image recovery for sampling not only a single output of reconstruction, but a variety of different outputs using a learned model of the posterior distribution. Sampling a multitude of image outputs makes it possible to evaluate the variance at the voxel-level to compute uncertainties within a given image. This technique of uncertainty quantification can be used to determine whether specific regions of interest have high uncertainty values around abnormal image findings.

For downstream image analysis tasks, techniques like Monte Carlo Dropout have gained popularity for estimating output uncertainty \cite{gal2016dropout}. Although dropout is typically employed for model regularization and reducing overfitting during training by randomly setting parameters to zero, it can also be applied at inference time. This involves producing multiple outputs with different neural network weights set to zero. This Monte Carlo Dropout approximates a Gaussian process, which facilitates the computation of uncertainty across the variance of all the generated inputs. Although straightforward to implement, this approach requires multiple forward passes during inference.

Looking forward, despite our ability to generate voxel-level uncertainties, the optimal utilization of these generated uncertainty maps still remains unclear. This elicits a number of intriguing research questions concerning the ways in which these uncertainties can be utilized beyond simply presenting them to end-users. For instance, it is unlikely that a radiologist would review both the output of a deep learning reconstruction and the corresponding uncertainty map since doing so would nearly double clinical read durations, and the radiologist may not necessarily know how to contextualize regions of high and low uncertainty. 

Alternatively, these uncertainty maps could be applied in iterative reconstruction techniques 
\cite{zhang2019reducing}. In this approach, the uncertainty associated with a specific step in an image reconstruction pipeline could serve as input for the subsequent step. The reconstruction network's objective would be to reconstruct the same image but with the new constraint of reducing the underlying uncertainty. This type of formulation holds potential for adaptive, case-based sampling schemes tailored to individual patients.

Another promising area of research involves directly integrating uncertainties from MR image reconstruction with those of the relevant downstream parameters of interest \cite{fischer2023uncertainty}. This methodology could serve to estimate maximum acceleration rates without compromising quantitative clinical parameters. Achieving such end-to-end analysis necessitates datasets that can provide raw case-based data alongside downstream image analysis datasets (e.g. SKM-TEA, fastMRI+, K2S, etc).

\section{Implementation issues }

Training DL reconstruction models can be computationally challenging, especially for large networks. For purposes of illustration, we focus on the case of unrolled networks.

\emph{\textbf{Memory demands of unrolled algorithms.}}
PnP and score-based algorithms that pre-train deep learning modules as denoisers are associated with low memory demand. In contrast, unrolled algorithms, which offer state-of-the art performance compared to PnP methods, involve a number of iterations and their training is restricted by the memory of the GPU devices during training. This often limits the applicability of unrolled algorithms to large-scale multi-dimensional problems. 

\emph{\textbf{Strategies for computational efficiency.}}
Several strategies have been introduced to overcome the memory limitations of unrolled methods.  For an unrolled network with $N$ iterations and shared CNN modules across iterations, the computational complexity and memory demands of backpropagation are $\mathcal O(N)$ and $\mathcal O(N)$, respectively. The forward steps can be recomputed during backpropagation, which reduces the memory demand to $\mathcal O(1)$, while the computational complexity increases to $\mathcal O(N^2)$. Forward checkpointing \cite{chen2016training} saves the variables for every $K$ layers during forward propagation, which reduces the computational demand to $\mathcal O(NK)$, while the memory demand is $\mathcal O(N/K)$. Reverse recalculation has been proposed to reduce the memory demand to $\mathcal O(1)$ and computational complexity to $\mathcal O(N)$ \cite{kellman2020memory}. However, the approach in \cite{kellman2020memory} requires multiple iterations to invert each CNN block, resulting in high computational complexity in practical applications. The deep equilibrium (DEQ) model \cite{bai2019deep} was recently adapted to inverse problems to significantly improve the memory demand \cite{deq,mol} of unrolled methods. Unlike unrolled methods, DEQ schemes run the iterations until convergence, similar to PnP algorithms. This property makes it possible to perform forward and backward propagation using fixed-point iteration involving a single physical layer, which reduces the memory demand to $\mathcal O(1)$, while the computational complexity is $\mathcal O(N)$; this offers better tradeoffs than the alternatives discussed above \cite{chen2016training,kellman2020memory}. The runtime of DEQ methods that are iterated until convergence are variable compared to unrolled methods, which use a finite number of iterations. In addition, the convergence of the iterative algorithm is crucial for the accuracy of backpropagation steps in DEQ, unlike in unrolled methods. Convergence guarantees were introduced in  \cite{deq,mol}.

\section{Conclusion}\label{sec13}

The introduction and rapid development of deep-learning-based strategies for MRI reconstruction have brought about a dramatic acceleration in acquisition time, paved the way for rapid and accurate parameter mapping, and facilitated automatic schedule optimization. While several key challenges still lie ahead, especially in terms of robust generalization, careful considerations of the training data diversity, ongoing model validation, and potentially, the development of adaptive or continuous learning systems are expected to enable adjustment to new data distributions over time.

% \bmhead{Acknowledgments}
\section*{Acknowledgments}

The authors thank Jonathan Tamir for useful comments regarding diffusion models. 
M.J acknowledges support from NIH grants R01 AG067078, R01EB019961, R01 EB031169, and Canon Medical Research.
O.P. acknowledges funding support from the Ministry of Innovation, Science and Technology, Israel, and the Tel Aviv University Center for AI and Data Science (TAD). E.S. is a Horev Fellow and acknowledges funding support from the Technion's Leaders in Science and Technology program. A.C. receives support from NIH grants R01 AR077604, R01 EB002524, and R01 AR079431; and from GE Healthcare, Philips, Amazon, Microsoft, and Stability AI.

\bibliography{DL_unified_updated_references,DL_few_extra_refs}

\end{document}